\documentclass{article}

\usepackage{microtype}
\usepackage{graphicx}
\usepackage{subfigure}
\usepackage{booktabs} 

\usepackage{hyperref}

\usepackage[accepted]{icml2025}

\usepackage{amsmath}
\usepackage{amssymb}
\usepackage{mathtools}
\usepackage{amsthm}
\usepackage{graphics}
\usepackage{epsfig}
\usepackage{multirow}
\usepackage{xspace}
\usepackage{subfigure}
\usepackage{fancyhdr}
\usepackage{microtype}
\usepackage{color}
\usepackage{colortbl}
\usepackage{xcolor}
\definecolor{mygray}{gray}{.85}
\definecolor{myyellow}{RGB}{204,102,0}
\definecolor{myred}{RGB}{204,0,102}
\definecolor{mypurple}{RGB}{102,0,204}
\definecolor{maroon}{cmyk}{0,0.87,0.68,0.32}
\definecolor{myblue}{RGB}{227,227,240}

\usepackage{graphics}
\usepackage{amssymb}
\usepackage{amsthm}
\usepackage{mathrsfs}
\usepackage{booktabs}
\usepackage{algorithm}
\usepackage{algorithmic}
\usepackage{arydshln}
\usepackage{siunitx}
\usepackage{appendix}
\usepackage{enumitem}
\usepackage{wrapfig}

\usepackage[capitalize,noabbrev]{cleveref}

\theoremstyle{plain}

\theoremstyle{definition}

\theoremstyle{remark}

\newcommand{\model}{PIDiff\xspace}
\newcommand{\dataset}{PIBench\xspace}
\newcommand{\network}{PIFormer\xspace}

\usepackage[textsize=tiny]{todonotes}

\icmltitlerunning{Joint Discrete and $SE(3)$ Diffusion for Protein Interface Design}
\begin{document}

\twocolumn[

\icmltitle{Joint Discrete and $SE(3)$ Diffusion for Protein Interface Design}


\icmlsetsymbol{equal}{*}

\begin{icmlauthorlist}
\icmlauthor{Zhenqiao Song}{yyy}
\icmlauthor{Jianwen Xie}{comp}
\icmlauthor{Lei Li}{yyy}
\end{icmlauthorlist}
\icmlaffiliation{yyy}{Language Technologies Institute, Carnegie Mellon University}
\icmlaffiliation{comp}{Lambda Inc}

\icmlcorrespondingauthor{Zhenqiao Song}{zhenqiaosong@cmu.edu}

\icmlkeywords{Machine Learning, ICML}

\vskip 0.3in
]



\printAffiliationsAndNotice{}  

\begin{abstract}
The design of high-affinity, target-specific protein binders is central to the development of modern therapeutics. However, existing protein-generation methods remain limited in this setting, as many generated designs fail to simultaneously achieve geometric and physicochemical complementarity at binding interfaces. To address this challenge, we introduce \model, a physicochemically enhanced multimodal diffusion framework for protein–protein complex design. \model jointly designs target-conditioned protein sequences and structures by coupling diffusion in discrete sequence space with diffusion on the continuous \(SE(3)\) structural manifold. At its core, \model employs \network, a novel architecture that interleaves geometry-aware invariant point attention with global sequence self-attention. We further curate \dataset, a  protein–protein complex dataset with diverse target scaffolds derived from the Protein Data Bank (PDB). We first pretrain \model on \dataset and subsequently fine-tune it for two therapeutically relevant applications: minibinder design and antibody design. PPFrame consistently outperforms baseline
methods across sequence design success rate, structure binding energy, and
sequence-structure co-designability on all tasks.

\end{abstract}

\section{Introduction}


The \textit{de novo} design of high-affinity, target-specific protein binders is foundational to modern biomedicine, driving transformative advances in therapeutics \citep{nelson2010development}, diagnostics \citep{brennan2010antibody}, and vaccines \citep{correia2014proof,sesterhenn2020novo}. Central to molecular recognition is the requirement for strict geometric and physicochemical complementarity across the binding interface \citep{chothia1975principles,conte1999atomic,gainza2023novo}. The complementary interface is governed by core energetic drivers: van der Waals forces optimize spatial packing, while hydrophobic and electrostatic interactions enforce physicochemical compatibility. 
Although traditional methods have exploited these energetic principles to engage pre-selected surface patches \citep{chevalier2017massively,silva2019novo}, they frequently fail when target proteins lack well-defined binding pockets, thereby restricting their scope to a narrow range of target topologies. 

Recently, rapid advances in deep generative modeling have significantly expanded binder design capabilities~\citep{evans2021protein,cao2022design,bennett2023improving,zambaldi2024novo,stark2025boltzgen,sappington2026improved}. These approaches have made it possible to design binders across diverse protein targets. In specialized domains, such as flexible peptides \citep{vazquez2024novo} and intrinsically disordered proteins \citep{wu2024sequence}, deep generative models can yield potent binders without requiring extensive experimental optimization. Despite these breakthroughs, the overall success rate remains low due to two primary failure modes: the designed structure may fail to bind target interfaces due to the physicochemical or geometric non-complementarity, and the designed sequence may not fold into the intended structure. Crucially, most existing frameworks decouple sequence and  structure generation or specialize in particular interaction types, leaving challenging, pocketless targets still intractable.

To address these challenges, we present \model, a physicochemically enhanced multimodal diffusion method to co-design binder protein sequence and structure for an arbitrary target. In our framework, proteins are represented as rigid-body frames, coupling each residue's spatial backbone coordinates with its local physicochemical properties. \model employs a unified multimodal diffusion process that operates jointly over discrete amino acid sequence space and the continuous $SE(3)$ structural manifold, progressively adding noise to train a joint generative model. The generative model is powered by \network, an architecture that interleaves invariant point attention \citep{jumper2021highly} with sequence self-attention \citep{vaswani2017attention} to simultaneously capture spatial geometric constraints and amino acid dependencies. By tightly integrating sequence, structure, and physicochemical properties, \model enforces binding interface complementarity and high sequence-structure consistency, effectively mitigating key failure modes in binder design to drive higher success rates.

Concretely, our contributions are listed as follows:
\begin{itemize}[nosep,leftmargin=2.0em]
\item We propose \model, a multimodal diffusion method building upon \network to co-design the binder protein sequences and structures for arbitrary protein targets.
\item We create \dataset, a general protein-protein complex dataset across diverse target geometries comprising 602,364 complexes curated from PDB~\citep{berman2000protein}. 
\item Our model pretrained on \dataset outperforms strong baselines, demonstrating a sequence design success rate of 47.8\%, structure binding energy of -67.8 and a co-design consistency of 65.2\%.
\item We finetune the pretrained \model on two important real-world applications: target-protein mini-binder complex design and antigen-antibody complex design. The finetuned models achieve design success rates of $40.7\%$ and $18.7\%$, respectively, demonstrating the framework's effectiveness in addressing key therapeutic challenges.
\end{itemize}

\section{Related Work}
\textbf{Protein-Protein Complex Design.}
Designing target-specific protein binders with high affinity has been extensively explored using a wide range of approaches. 
Conventional experimental approaches rely primarily on directed evolution of natural protein scaffolds \citep{vazquez2015challenges} or high-throughput antibody screening from synthetic libraries \citep{chao2006isolating,hackel2008picomolar}.  Although powerful, these methods require considerable experimental effort and do not provide substantial control over the properties of the binder proteins.
Traditional computational methods have sought to model physicochemical properties to accelerate the design process \citep{chevalier2017massively,fleishman2011computational}.
However, their reliance on pre-defined hotspot residues heavily restricts their scope. Recent advances in deep learning have significantly improved the prediction and design of protein-protein interactions~\citep{evans2021protein,humphreys2021computed,bryant2022improved,pacesa2024bindcraft,didi2026scaling}, enabling binder design without prior hotspot specification~\citep{cao2022design,bennett2023improving,chen2025all}. 
Nevertheless, the overall success rate remains low due to the difficulty of simultaneously satisfying the strict geometric and physicochemical complementarity essential for stable physical binding interfaces~\citep{berger2024preclinical,zambaldi2024novo}.

\textbf{Diffusion Models.} Diffusion models \citep{sohl2015deep} have established themselves as a powerful class of generative frameworks. \citet{ho2020denoising} introduce denoising diffusion probabilistic models, formalizing their connection to score-based generative modeling \citep{song2019generative}. Diffusion models have achieved remarkable success in computational protein design. Early structural approaches focus on backbone design, progressing from $C_\alpha$ trace generation \citep{trippe2022diffusion} to full rigid-body $SE(3)$ backbone modeling \citep{watson2023novo,yim2023se}. In parallel, discrete diffusion formulations are introduced to tackle protein sequence generation \citep{alamdari2023protein,wang2024diffusion}. More recent efforts have explored joint sequence-structure co-design through multimodal diffusion~\citep{song2025ppdiff,wang2025dplm}. However, the application of diffusion models for the sequence-structure co-design of protein complexes still remains an open problem.

\section{Proposed Method: \model}
\subsection{Problem Definition}
\label{Sec:problem_definition}
We model a protein as a joint representation of its amino acid sequence and 3D backbone structure. Let $\mathcal{A}$ denote the set of $20$ standard amino acids. A protein of length $L$ is represented by a sequence $\boldsymbol{a}=\{a_1, a_2, ..., a_L\} \in \mathcal{A}^L$ and an associated backbone structure $\boldsymbol{s}$. Following \citet{jumper2021highly}, the backbone geometry is parameterized by a collection of local rigid frames mapping from reference coordinates $[N^*, C_{\alpha}^*, C^*, O^*] \in \mathbb{R}^{4 \times 3}$ centered at $C_{\alpha}^* = (0, 0, 0)$. Specifically, for residue $i \in \{1, \dots, L\}$, the mainchain heavy atom coordinates $[N_i, C_i, (C_{\alpha})_i]$ are represented by applying a rigid body transformation $\boldsymbol{T}_i \in SE(3)$ to the reference frame: $[N_i, C_i, (C_{\alpha})_i] = \boldsymbol{T}_i \cdot [N^*, C^*, C_{\alpha}^*]$. Each frame $\boldsymbol{T}_i = (\boldsymbol{r}_i, \boldsymbol{x}_i)$ comprises a rotation matrix $\boldsymbol{r}_i \in SO(3)$ and a translation vector $\boldsymbol{x}_i \in \mathbb{R}^3$, acting on a spatial coordinate $\boldsymbol{v} \in \mathbb{R}^3$ via $\boldsymbol{T}_i \cdot \boldsymbol{v} = \boldsymbol{r}_i \boldsymbol{v} + \boldsymbol{x}_i$. The full backbone structure is thus defined as $\boldsymbol{s} = \{\boldsymbol{T}_1 \cdot [N^*, C^*, C_{\alpha}^*], \dots, \boldsymbol{T}_L\cdot [N^*, C^*, C_{\alpha}^*]\}$, with backbone oxygen atoms positioned using the torsion angle $\psi$ around the $C_{\alpha}\text{--}C$ bond. Furthermore, each residue $i$ is assigned a physicochemical feature vector $\boldsymbol{f}_i = (y_i, c_i)$ encoding its hydrophobicity $y_i$ and charge $c_i$, yielding a protein feature sequence $\boldsymbol{f} = (\boldsymbol{f}_1, \boldsymbol{f}_2, \dots, \boldsymbol{f}_L)$.

Given a target protein $\mathcal{T} = \{\boldsymbol{a}^{\mathcal{T}}, \boldsymbol{s}^{\mathcal{T}}, \boldsymbol{f}^{\mathcal{T}}\}$ of length $M$, the objective of protein-protein complex co-design is to generate a binder protein $\mathcal{B} = \{\boldsymbol{a}^{\mathcal{B}}, \boldsymbol{s}^{\mathcal{B}}\}$ of length $N$ displaying complementary physicochemical and geometric properties to target $\mathcal{T}$. This entails jointly sampling the binder sequence $\boldsymbol{a}^{\mathcal{B}} = \{a_1^{\mathcal{B}}, \dots, a_N^{\mathcal{B}}\}$ and backbone transformations $\boldsymbol{T}^{\mathcal{B}} = \{\boldsymbol{T}_1^{\mathcal{B}}, \dots, \boldsymbol{T}_N^{\mathcal{B}}\}$ with aligned physicochemical features $\boldsymbol{f}^{\mathcal{B}}$. Formally, we formulate this co-design task as learning a conditional generative model $p_{\theta}(\mathcal{B} \mid \mathcal{T})$ parameterized by $\theta$.

\begin{figure*}
  \centering
  \includegraphics[width=12cm]{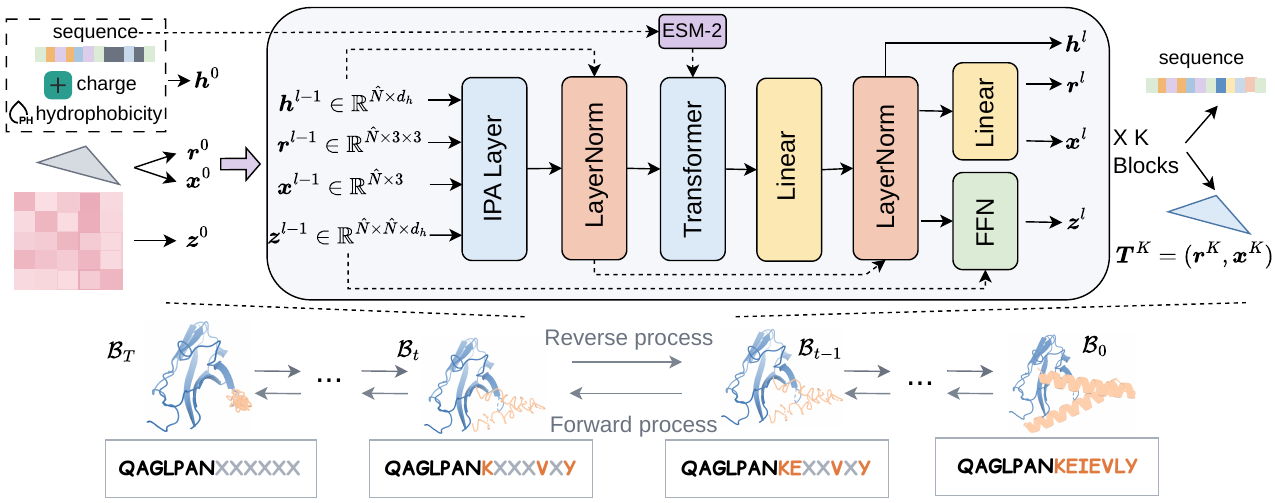}
  \vspace{-1.0em}
  \caption{
  Overview of the proposed \model architecture. At each diffusion time step, the target protein structure is rendered in green, while the binder protein is shown in pink.}
  \label{Fig: model}
\end{figure*}

\subsection{Overall Model Architecture}
\label{Sec:overall_architecture}
The overall architecture of \model is illustrated in Figure~\ref{Fig: model}.
\model is a multimodal denoising diffusion probabilistic model (DDPM)~\citep{ho2020denoising} designed for joint generation of a binder's primary sequence $\boldsymbol{a}^{\mathcal{B}}$ and backbone structure $\boldsymbol{s}^{\mathcal{B}}$ conditioned on a target protein $\mathcal{T}$. The framework models a forward perturbation process $q$ that corrupts data state $\mathcal{B}_0 \sim q(\mathcal{B}_0\mid\mathcal{T})$ into noise $\mathcal{B}_\mathbb{T} \sim q_{\text{noise}}$ over $\mathbb{T}$ steps, alongside a parameterized reverse generative process $p_{\theta}$:
\begin{equation}
\small  
\label{Eq:diffusion_processes}
\begin{split}
q(\mathcal{B}_{1:\mathbb{T}}\mid\mathcal{B}_0,\mathcal{T}) &= \prod\nolimits_{t=1}^\mathbb{T} q(\mathcal{B}_t\mid\mathcal{B}_{t-1},\mathcal{T})\\
p_{\theta}(\mathcal{B}_{0:\mathbb{T}}\mid\mathcal{T}) &= p(\mathcal{B}_\mathbb{T})\prod\nolimits_{t=1}^\mathbb{T} p_{\theta}(\mathcal{B}_{t-1}\mid\mathcal{B}_t, \mathcal{T})
\end{split}
\end{equation}
Each latent variable $\mathcal{B}_t = (\boldsymbol{a}^{\mathcal{B}}_t, \boldsymbol{s}^{\mathcal{B}}_t)$ represents a hybrid state spanning discrete amino acid categories and continuous $SE(3)$ transformations. Model parameters $\theta$ are optimized by maximizing the standard evidence lower bound on log-likelihood:
\begin{equation}
\small 
\label{Eq:diffusion_loss}
\begin{split}
\mathcal{L}_{\text{VLB}} &= \mathbb{E}_{q}\Big[ \log p_{\theta}(\mathcal{B}_0\mid\mathcal{B}_1,\mathcal{T}) \\&- \sum\nolimits_{t=2}^\mathbb{T} \underbrace{\mathrm{KL}\left[ q(\mathcal{B}_{t-1}\mid\mathcal{B}_t, \mathcal{B}_0, \mathcal{T}) \parallel p_{\theta}(\mathcal{B}_{t-1}\mid\mathcal{B}_t, \mathcal{T}) \right]}_{\mathcal{L}_t}  \Big]
\end{split}
\end{equation}
By unifying discrete sequence transitions with continuous manifold diffusion, \model enhances sequence-structure consistency throughout generative sampling.

\subsection{\model Diffusion Process}
\label{Sec:diffusion_process}
We formulate the joint forward diffusion process of protein residue sequence $\boldsymbol{a}_t^{\mathcal{B}}$ and backbone structure $\boldsymbol{s}^{\mathcal{B}}_t$ as factorized transitions across individual residues and modalities at each time step $t$:
\begin{equation}
\small 
\label{Eq:diffusion_process_time_step}
\begin{split}
q(\mathcal{B}_t \mid \mathcal{B}_{t-1}, \mathcal{T}) &= \prod\nolimits_{i=1}^N q(a^{\mathcal{B}}_{i,t} \mid a^{\mathcal{B}}_{i,t-1}, \mathcal{T}) \\&\cdot \prod\nolimits_{i=1}^N q(\boldsymbol{s}^{\mathcal{B}}_{i,t} \mid \boldsymbol{s}^{\mathcal{B}}_{i,t-1}, \mathcal{T})
\end{split}
\end{equation}
where $a_{i,t}^{\mathcal{B}}$ and $\boldsymbol{s}_{i,t}^{\mathcal{B}} = \boldsymbol{T}_{i,t}^{\mathcal{B}} \cdot [N^*, C_{\alpha}^*, C^*]$ denote the $i$-th residue type and local rigid frame at time step $t$, respectively. This factorization enables efficient noise sampling across different modalities.

\noindent\textbf{Discrete Sequence Diffusion.} For the sequence components $\boldsymbol{a}^{\mathcal{B}}$, state transitions are defined via transition matrices $[\boldsymbol{Q}_t]_{mn} = q(a^{\mathcal{B}}_{i,t} = n \mid a^{\mathcal{B}}_{i,t-1} = m)$. Expressing $a_{i,t}^{\mathcal{B}}$ as a one-hot row vector $\boldsymbol{a}_{i,t}^{\mathcal{B}}$, the single-step transition is given by:
\begin{equation}
\small 
q(\boldsymbol{a}^{\mathcal{B}}_{i,t} \mid \boldsymbol{a}^{\mathcal{B}}_{i,t-1}, \mathcal{T}) = \operatorname{Cat}(\boldsymbol{a}^{\mathcal{B}}_{i,t}; \boldsymbol{p} = \boldsymbol{a}^{\mathcal{B}}_{i,t-1}\boldsymbol{Q}_t)
\end{equation}
where $\operatorname{Cat}(\boldsymbol{a}; \boldsymbol{p})$ denotes a categorical distribution with probability vector $\boldsymbol{p}$. Starting from $\boldsymbol{a}^{\mathcal{B}}_{i,0}$, the $t$-step marginal and posterior distributions are analytically derived as:
\begin{equation}
\small 
\begin{split}
&q(\boldsymbol{a}^{\mathcal{B}}_{i,t} \mid \boldsymbol{a}^{\mathcal{B}}_{i,0}, \mathcal{T}) = \operatorname{Cat}(\boldsymbol{a}^{\mathcal{B}}_{i,t}; \boldsymbol{p} = \boldsymbol{a}^{\mathcal{B}}_{i,0}\bar{\boldsymbol{Q}}_t), \quad \quad \bar{\boldsymbol{Q}}_t = \prod\nolimits_{j=1}^t \boldsymbol{Q}_j \\
&q(\boldsymbol{a}^{\mathcal{B}}_{i,t-1}|\boldsymbol{a}^{\mathcal{B}}_{i,t}, \boldsymbol{a}^{\mathcal{B}}_{i,0}, \mathcal{T})=\text{Cat}\Big (\boldsymbol{a}^{\mathcal{B}}_{i,t-1};\boldsymbol{p}=\frac{\boldsymbol{a}^{\mathcal{B}}_{i,t}\boldsymbol{Q}_t^\top \odot \boldsymbol{a}^{\mathcal{B}}_{i,0}\boldsymbol{\overline{Q}}_{t-1}}{\boldsymbol{a}^{\mathcal{B}}_{i,0}\boldsymbol{\overline{Q}}_t {\boldsymbol{a}^{\mathcal{B}\top}_{i,t}}}\Big )\\
\end{split}
\end{equation}
Following \citet{austin2021structured}, we adopt an absorbing-state transition matrix $\boldsymbol{Q}_t$ with an absorbing token [mask], where each token either remains unchanged or transitions to [mask] with probability $\beta_t$:
\begin{equation}
\small 
[\boldsymbol{Q}_t]_{mn} = \begin{cases}
1 & \text{if } m = n = \text{[mask]} \\ 
1 - \beta_t & \text{if } m = n \neq \text{[mask]} \\
\beta_t & \text{if } n = \text{[mask]}, m \neq \text{[mask]}
\end{cases}
\end{equation}

\noindent\textbf{Continuous Structure Diffusion on $SE(3)$.} To generate backbone structures in $SE(3)^N$, we build upon Riemannian score-based generative modeling \citep{de2022riemannian}. Following \citet{yim2023se}, we treat the rotational $SO(3)$ and translational $\mathbb{R}^3$ sub-manifolds independently, setting the final invariant distribution to $p_{\mathrm{SE}(3)}(\boldsymbol{T}^{\mathcal{B}}_{i,t}) \propto \mathcal{U}_{\mathrm{SO}(3)}(\boldsymbol{r}^{\mathcal{B}}_{i,t}) \, \mathcal{N}(\boldsymbol{x}^{\mathcal{B}}_{i,t}; \mathbf{0}, \boldsymbol{I}_3)$. The forward Stochastic Differential Equation (SDE) on $\boldsymbol{T}^{\mathcal{B}}_{i,t} = (\boldsymbol{r}^{\mathcal{B}}_{i,t}, \boldsymbol{x}^{\mathcal{B}}_{i,t})$ is defined by:
\begin{equation}
\small 
d\boldsymbol{T}^{\mathcal{B}}_{i,t} = \left[ \mathbf{0}, -\frac{1}{2}\boldsymbol{x}^{\mathcal{B}}_{i,t} \right] dt + \left[ d\boldsymbol{B}^{(i,t)}_{\mathrm{SO}(3)}, d\boldsymbol{B}^{(i,t)}_{\mathbb{R}^3} \right]
\end{equation}
where $\boldsymbol{B}_{\mathrm{SO}(3)}$ and $\boldsymbol{B}_{\mathbb{R}^3}$ denote Brownian motions on $SO(3)$ and $\mathbb{R}^3$, respectively. The score network $s_{\theta}(t, \boldsymbol{T}_{i,t}^{\mathcal{B}}; \mathcal{T})$ is trained via denoising score matching:
\begin{equation}
\small 
\label{Eq:dsm_loss}
\mathcal{L}_{\text{struct}} = \mathbb{E}\left[ \lambda_t \left\| \nabla_{\boldsymbol{T}^{\mathcal{B}}_{i,t}}\log p_{t\mid 0}(\boldsymbol{T}^{\mathcal{B}}_{i,t} \mid \boldsymbol{T}^{\mathcal{B}}_{i,0}; \mathcal{T}) - s_{\theta}(t, \boldsymbol{T}_{i,t}^{\mathcal{B}}; \mathcal{T}) \right\|^2 \right]
\end{equation}
By leveraging independence across coordinates, the conditional score decomposes into rotational and translational components: $\nabla_{\boldsymbol{T}^{\mathcal{B}}_{i,t}}\log p_{t\mid 0} = [\nabla_{\boldsymbol{r}^{\mathcal{B}}_{i,t}}\log p_{t\mid 0}, \, \nabla_{\boldsymbol{x}^{\mathcal{B}}_{i,t}}\log p_{t\mid 0}]$. For $SO(3)$, the perturbation kernel $p_{t\mid 0}$ is governed by the heat kernel on $SO(3)$ \citep{hsu2002stochastic}, enabling analytical score evaluation:
\begin{equation}
\small 
\begin{split}
&\nabla_{\boldsymbol{r}^{\mathcal{B}}_{i,t}}\log p_{t\mid 0}(\boldsymbol{r}^{\mathcal{B}}_{i,t} \mid \boldsymbol{r}^{\mathcal{B}}_{i,0}, \mathcal{T}) = \frac{\boldsymbol{r}^{\mathcal{B}}_{i,t}}{\omega_{i,t}} \log\left( (\boldsymbol{r}^{\mathcal{B}}_{i,0})^\top \boldsymbol{r}^{\mathcal{B}}_{i,t} \right) \frac{\partial_{\omega} f(\omega_{i,t}, t)}{f(\omega_{i,t}, t)} \\ 
&f(\omega, t) = \sum_{l=0}^{\infty} (2l+1) e^{-l(l+1)t/2} \frac{\sin((l+1/2)\omega)}{\sin(\omega/2)}
\end{split}
\end{equation}
where $\omega_{i,t} = \omega((\boldsymbol{r}^{\mathcal{B}}_{i,0})^\top \boldsymbol{r}^{\mathcal{B}}_{i,t})$ is the rotation angle in radians. For $\mathbb{R}^3$, the translation process follows an Ornstein-Uhlenbeck process \citep{song2020score} with kernel $p_{t\mid 0}(\boldsymbol{x}^{\mathcal{B}}_{i,t} \mid \boldsymbol{x}^{\mathcal{B}}_{i,0}) = \mathcal{N}(\boldsymbol{x}^{\mathcal{B}}_{i,t}; e^{-t/2}\boldsymbol{x}^{\mathcal{B}}_{i,0}, (1-e^{-t})\boldsymbol{I}_3)$, yielding the exact conditional score:
\begin{equation}
\small 
\nabla_{\boldsymbol{x}^{\mathcal{B}}_{i,t}}\log p_{t\mid 0}(\boldsymbol{x}^{\mathcal{B}}_{i,t} \mid \boldsymbol{x}^{\mathcal{B}}_{i,0}, \mathcal{T}) = (1 - e^{-t})^{-1} \left( e^{-t/2}\boldsymbol{x}^{\mathcal{B}}_{i,0} - \boldsymbol{x}^{\mathcal{B}}_{i,t} \right)
\end{equation}

\subsection{\network Architecture for the Reverse Generative Process}
\label{Sec:network_architecture}
The reverse generative process parameterizes the conditional transition distribution $p_{\theta}(\mathcal{B}_{t-1} \mid \mathcal{B}_t, \mathcal{T})$ to reconstruct uncorrupted sequence-structure pairs from noisy states. We parameterize this reverse mapping using \network, a multi-block neural architecture designed for joint sequence-structure co-design. \network iteratively refines spatial frames and sequence representations across $K$ stacked blocks. Each block interleaves spatial-based invariant point attention~(IPA) layer \citep{jumper2021highly} with sequence-based self-attention layers, effectively modeling 3D geometric interactions alongside global amino acid dependencies.

Node embeddings $\boldsymbol{h}^0 = (\boldsymbol{h}^0_1, \dots, \boldsymbol{h}^0_{\hat{N}}) \in \mathbb{R}^{\hat{N} \times d_h}$ are initialized with residue-level physicochemical features, while edge embeddings $\boldsymbol{z}^0 \in \mathbb{R}^{\hat{N} \times \hat{N} \times d_z}$ capture pairwise initial geometry following \citet{trippe2022diffusion}, where $\hat{N} = M + N$ denotes the total residue length of the target-binder complex. To enhance sequence modeling capacity, we incorporate pretrained residue representations $\boldsymbol{e} \in \mathbb{R}^{\hat{N} \times d_e}$ extracted from ESM-2 \citep{lin2022language}. Within the $l$-th block ($l \in \{1, \dots, K\}$), representations and rigid frames are updated as follows:
\begin{equation}
\small 
\label{Eq:network_layer_update}
\begin{aligned}
\boldsymbol{h}^{(l-0.75)} &= \mathrm{IPALayer}\big(\boldsymbol{h}^{(l-1)}, \boldsymbol{z}^{(l-1)}, \boldsymbol{T}^{(l-1)}\big), \\
\boldsymbol{h}^{(l-0.5)} &= \mathrm{LayerNorm}\big(\boldsymbol{h}^{(l-0.75)} + \boldsymbol{h}^{(l-1)}\big) \\
\boldsymbol{h}^{(l-0.25)} &= \mathrm{Transformer}\big(\boldsymbol{h}^{(l-0.5)} + \boldsymbol{e}\big) \\
\boldsymbol{h}^l &= \mathrm{LayerNorm}\big(\mathrm{Linear}(\boldsymbol{h}^{(l-0.25)}) + \boldsymbol{h}^{(l-0.5)}\big) \\
\tilde{\boldsymbol{T}}^l &= \boldsymbol{h}^l \boldsymbol{W}_f \\ 
\boldsymbol{r}^l &= \boldsymbol{r}^{l-1} + \boldsymbol{r}^{l-1} \cdot \tilde{\boldsymbol{T}}^l_{[:3]}, \boldsymbol{x}^l = \boldsymbol{x}^{l-1} + \boldsymbol{r}^{l-1} \cdot \tilde{\boldsymbol{T}}^l_{[3:]}\\ 
\boldsymbol{T}^l &= (\boldsymbol{r}^l, \boldsymbol{x}^l) \\
\boldsymbol{z}^l&=\text{FFN}\big(f(\boldsymbol{h}^l, \boldsymbol{z}^{(l-1)})\big)\\ f_{ij}(\boldsymbol{h}^l, \boldsymbol{z}^{(l-1)})&=\text{Concatenate}\big(\text{Linear}(\boldsymbol{h}^l_i), \text{Linear}(\boldsymbol{h}^l_j), \boldsymbol{z}^{(l-1)}_{ij}\big)
\end{aligned}
\end{equation}
where $\boldsymbol{W}_f \in \mathbb{R}^{d_h \times 6}$ is a trainable projection matrix, and $\boldsymbol{T}^0 = [\boldsymbol{T}^{\mathcal{T}}, \tilde{\boldsymbol{T}}^{\mathcal{B}}]$ concatenates ground-truth target frames $\boldsymbol{T}^{\mathcal{T}}$ with perturbed binder frames $\tilde{\boldsymbol{T}}^{\mathcal{B}}$. 
FFN denotes the feed-forward layer.

From the final outputs $(\boldsymbol{h}^K, \boldsymbol{T}^K)$, we compute rotational and translational scores, predict the uncorrupted sequence distribution, and estimate backbone $\psi$ torsion angles:
\begin{equation}
\small 
\label{Eq:predictions}
\begin{split}
s^{r,i}_{\theta} &= f_r(\tilde{\boldsymbol{r}}_i^{\mathcal{B}}, \boldsymbol{r}_i^{\mathcal{B}, K}), \quad
s^{x,i}_{\theta} = f_x(\tilde{\boldsymbol{x}}_i^{\mathcal{B}}, \boldsymbol{x}_i^{\mathcal{B}, K}) \\
p_{\theta}(\hat{\boldsymbol{a}}_{i,0}^{\mathcal{B}}) &= \mathrm{Softmax}\big(\mathrm{Linear}(\boldsymbol{h}_i^K)\big), \quad 
\psi_i = f_{\psi}(\boldsymbol{h}_i^K)
\end{split}
\end{equation}
where $f_r$ and $f_x$ compute score function approximations from noisy frames $(\tilde{\boldsymbol{r}}_i^{\mathcal{B}}, \tilde{\boldsymbol{x}}_i^{\mathcal{B}})$ and network-predicted frames $(\boldsymbol{r}_i^{\mathcal{B}, K}, \boldsymbol{x}_i^{\mathcal{B}, K})$, and $f_{\psi}$ maps node representations to $\psi$ torsion angles. The predicted distribution $p_{\theta}(\hat{\boldsymbol{a}}_{i,0}^{\mathcal{B}})$ parameterizes the clean sequence $\boldsymbol{a}_{i,0}^{\mathcal{B}}$, allowing analytical calculation of the reverse categorical posterior $p_{\theta}(\boldsymbol{a}^{\mathcal{B}}_{i,t-1}|\boldsymbol{a}^{\mathcal{B}}_{i,t}, \hat{\boldsymbol{a}}^{\mathcal{B}}_{i,0}, \mathcal{T})$.

\subsection{Training Objective}
\label{Sec:training_objective}
To achieve effective sequence-structure co-design, \model is optimized via a multitask learning objective $\mathcal{L}_t$ that combines discrete sequence learning, continuous $SE(3)$ score-matching frame losses, and late-stage auxiliary physical regularization with weights $\alpha$, $\gamma$, and $\omega$:
\begin{equation}
\small 
\label{Eq:total_loss}
\begin{split}
\mathcal{L}_t &= \alpha \mathcal{L}_{\text{seq}} + \gamma (\mathcal{L}_{\text{struct}}^r + \mathcal{L}_{\text{struct}}^x) + \omega \mathcal{L}_{\text{aux}}\\
\mathcal{L}_{\text{seq}} &= \sum\nolimits_{i=1}^N \mathrm{KL}\left( q(\boldsymbol{a}^{\mathcal{B}}_{i,t-1} \mid \boldsymbol{a}^{\mathcal{B}}_{i,t}, \boldsymbol{a}^{\mathcal{B}}_{i,0}, \mathcal{T}) \;\parallel\; p_{\theta}(\boldsymbol{a}^{\mathcal{B}}_{i,t-1}|\boldsymbol{a}^{\mathcal{B}}_{i,t}, \hat{\boldsymbol{a}}^{\mathcal{B}}_{i,0}, \mathcal{T}) \right) \\
\mathcal{L}_{\text{struct}}^r &= \sum\nolimits_{i=1}^N \left\| \nabla_{\boldsymbol{r}_{i,t}^{\mathcal{B}}} \log p_{t\mid 0}(\boldsymbol{r}_{i,t}^{\mathcal{B}} \mid \boldsymbol{r}_{i,0}^{\mathcal{B}}, \mathcal{T}) - s_{\theta}^{r,i} \right\|^2, \\ 
\mathcal{L}_{\text{struct}}^x &= \sum\nolimits_{i=1}^N \left\| \boldsymbol{x}_i^{\mathcal{B}} - \boldsymbol{x}_i^{\mathcal{B},K} \right\|^2
\end{split}
\end{equation}
where $\mathcal{L}_{\text{seq}}$ maximizes the variational lower bound in Eq.~\ref{Eq:diffusion_loss}, and the frame losses minimize the score-matching objective in Eq.~\ref{Eq:dsm_loss}. By applying target weighting schedules $\lambda_{i,t}^r = 1 / \mathbb{E}\big[\Vert\nabla \log p_{t\mid 0}(\boldsymbol{r}_{i,t} \mid \boldsymbol{r}_{i,0})\Vert_{\mathrm{SO}(3)}^2\big]$ for rotations and $\lambda_{i,t}^x = (1 - e^{-t}) / e^{-t/2}$ for translations, the frame objectives simplify to the unweighted formulations above. 

To enforce physical contact fidelity during final structure generation, we activate two auxiliary terms, $\mathcal{L}_{\text{aux}} = \mathcal{L}_{\text{struct}}^{\text{atom}} + \mathcal{L}_{\text{struct}}^{\text{pair}}$, during late denoising time steps ($t < 0.25$):
\begin{equation}
\small 
\label{Eq:aux_loss}
\begin{aligned}
\mathcal{L}_{\text{struct}}^{\text{atom}} &= \frac{1}{4N} \sum_{i=1}^N \sum_{m \in \mathbb{A}} \left\| \boldsymbol{x}_{i,m}^{\mathcal{B}} - \boldsymbol{x}^{\mathcal{B},K}_{i,m} \right\|^2, \quad \mathbb{A} = \{\mathrm{N}, \mathrm{C}_{\alpha}, \mathrm{C}, \mathrm{O}\} \\
\mathcal{L}_{\text{struct}}^{\text{pair}} &= \frac{1}{S} \sum_{1 \le i \neq j \le N} \sum_{m,n \in \mathbb{A}} \mathbb{I}(d^{ij}_{mn} < \tau) \left\| d^{ij}_{mn} - \hat{d}^{ij}_{mn} \right\|^2 \\ S &= \sum_{1 \le i \neq j \le N} \sum_{m,n \in \mathbb{A}} \mathbb{I}(d^{ij}_{mn} < \tau)
\end{aligned}
\end{equation}
where $d^*$ denotes the pairwise atomic distances, and $\tau$ specifies the interaction distance cutoff.

\subsection{Self-Conditioned Inference}
\label{Sec:self_conditioned_inference}
During inference, conditioned on the target protein's sequence $\boldsymbol{a}^{\mathcal{T}}$, structure $\boldsymbol{s}^{\mathcal{T}}$, and physicochemical features $\boldsymbol{f}^{\mathcal{T}}$, \model conditionally co-samples the binder sequence $\boldsymbol{a}^{\mathcal{B}}$ and backbone structure $\boldsymbol{s}^{\mathcal{B}}$ to yield high-affinity complexes. Sequence generation initializes all binder positions in an absorbing state [mask], fixing residue types and updating physicochemical features $\boldsymbol{f}^{\mathcal{B}}$ upon state transition. Structure generation utilizes Euler-Maruyama geodesic random walks on $SE(3)$ over $\mathbb{T}$ steps \citep{de2022riemannian}. Translations receive Gaussian updates from $\mathcal{N}(\mathbf{0}, \boldsymbol{I}_3)$, while rotations update along the Lie algebra $SO(3)$ tangent space as $\boldsymbol{r}'_{i,t} = \boldsymbol{r}_{i,t} \sum_{j=1}^3 \delta_j \boldsymbol{e}_j$ with $\delta_j \stackrel{\text{i.i.d.}}{\sim} \mathcal{N}(0, 1)$ and $\boldsymbol{e}_1, \boldsymbol{e}_2, \boldsymbol{e}_3$ are orthonormal basis vectors.
To maximize structural fidelity, we implement a self-conditioned inference scheme at each time step $t$: \network first performs an intermediate forward pass to update $C_{\alpha}$ positions, then recalculates spatial pair embeddings to refine sequence likelihoods and rigid frames in a second pass.

\section{Experiments}

\begin{table*}[!t]
\scriptsize 
\begin{center}
\setlength{\tabcolsep}{1.6mm}{
\begin{tabular}{lcccccccc}
\toprule
\multirow{2}{*}{Models} & \multicolumn{5}{c}{Sequence } & \multicolumn{2}{c}{Structure } & \multirow{2}{*}{Co-designability \scriptsize{($\uparrow$)}}  \\
\cmidrule(r){2-6} \cmidrule(r){7-8} 
& ipTM \scriptsize{($\uparrow$)}	& pLDDT \scriptsize{($\uparrow$)}	& SR \scriptsize{($\uparrow$)}	& Diversity \scriptsize{($\uparrow$)} &	Novelty \scriptsize{($\uparrow$)}&	$\Delta$G \scriptsize{($\downarrow$)}	& Designability \scriptsize{($\uparrow$)} &  \\
\midrule
RFD+IF&0.523&	81.61&	21.73\%&	37.90\%&	73.01\%&	-45.97&	50.00\%	&50.00\%\\
APM&0.252&	77.66&	2.17\%&	31.26\%&	53.27\%&	-45.12&	23.91\%&	34.78\%\\
Complexa&0.521&	84.61&	30.43\%&	\textbf{53.77\%}&	\textbf{82.69\%}&-56.57&56.52\%	&54.34\%\\
\model &\textbf{0.592}&	\textbf{87.18}&	\textbf{47.83\%}&	33.22\%&	57.49\%&	\textbf{-67.80}&	\textbf{65.22\%}	&\textbf{65.22\%}\\
\bottomrule
\end{tabular}}
\end{center}
\caption{Model performance on general protein-protein complex design task. 
Our \model achieves the best sequence design success rate, structure binding energy and sequence-structure co-designability.}
\label{Table:general_protein_protein_complex_design_performance}
\end{table*}

\subsection{\dataset Construction}
\label{Sec:dataset_curation}
To establish a foundation model for protein-protein complex design that can be seamlessly adapted to real-world applications, we curate a comprehensive dataset called \dataset. Following the data processing pipeline of AlphaFold3 \citep{abramson2024accurate}, we extract 301,182 chain-pair interfaces from the PDB. Complete processing workflows are detailed in Appendix~\ref{appendix:dataset_curation_details}.
By treating both chains within each interface symmetrically as target-binder pairs, we construct a benchmark comprising 602,364 complexes across diverse scaffolds. To prevent data leakage, sequences are clustered at 50\% identity, yielding 28,884 distinct clusters. We reserve 10 clusters each for validation and testing, allocating the remaining clusters for training. Detailed data statistics are given in Appendix Table~\ref{Tab: ppibench_data_statistics}.

\subsection{Experimental Setup}
\label{Sec:experimental_setup}
\noindent\textbf{Implementation Details.}
The \network comprises five blocks, each containing one IPA layer and six self-attention layers. Parameters for the self-attention layers are initialized from the pretrained 150M ESM-2 model. Node and edge embedding dimensions are set to 640 and 320, respectively, bringing the total parameter count of \model to 240M. The diffusion process is configured for 1,000 steps. In inference, for each target protein, we generate 10 binder candidates and rank them by their complex pLDDT scores, selecting the top-$k$ candidates. More implementation details are provided in Appendix~\ref{appendix:implementation_details}.

\noindent\textbf{Baseline Models.}
We compare the proposed \model against the following representative protein complex co-design methods:
(1) \textbf{RFD}+\textbf{ProteinMPNN (IF)} first applies RFDiffusion~\citep{watson2023novo} to design a binder protein structure based on the given target and then uses ProteinMPNN~\citep{dauparas2022robust} to generate the corresponding sequence.
(2) \textbf{APM} \citep{chen2025all} is an all-atom generative model for protein complex design.
(3) \textbf{Proteína-Complexa} \citep{didi2026scaling} is a fully atomistic binder generation method showing strong performance on protein-binding protein design. 

\noindent\textbf{Evaluation Metrics.}
To comprehensively evaluate \model, we assess generated complexes across three dimensions: sequence quality, structure reliability, and sequence-structure consistency. \textbf{Sequence Evaluation:} We evaluate designed sequences using five metrics: (1) \textbf{ipTM} via AlphaFold3 (AF3) to measure interface confidence; (2) \textbf{pLDDT} via AF3 to assess complex structure stability; (3) \textbf{Success Rate (SR)} defined as the fraction of designed candidates that simultaneously achieve an $\mathrm{ipTM} > 0.8$ and a $\mathrm{pLDDT} > 80$;
(4) \textbf{Diversity}, measured as the amino acid variation among candidates for the same target; and (5) \textbf{Novelty}, calculated as the sequence dissimilarity between candidates and reference sequences.
\textbf{Structure Evaluation:} We assess designed structures via: (6) \textbf{Binding Energy ($\Delta G$)}, computed via Rosetta \citep{chaudhury2010pyrosetta} to estimate interface energetic stability, where lower values indicate stronger binding propensity, and (7) \textbf{Designability}, defined as the fraction of designs achieving a self-consistency RMSD ($\mathrm{scRMSD}$) $< 2\,\text{\AA}$ \citep{trippe2022diffusion}.
\textbf{Sequence-Structure Consistency:} We measure joint consistency via (8) \textbf{Co-designability (Consistency)}, defined as the proportion of generated binder sequences whose AF3-folded structure matches the corresponding designed structures within $2\,\text{\AA}$ RMSD \citep{didi2026scaling}.

\begin{table*}[!t]
\scriptsize 
\begin{center}
\setlength{\tabcolsep}{1.0mm}{
\begin{tabular}{lccccccccc}
\toprule
\multirow{2}{*}{Models} & \multicolumn{3}{c}{FGFR2 (Seen) } & \multicolumn{3}{c}{InsulinR (Seen) } & \multicolumn{3}{c}{PDGFR (Seen)}  \\
\cmidrule(r){2-4} \cmidrule(r){5-7} \cmidrule(r){8-10}
& SR \scriptsize{($\uparrow$)}&	$\Delta$G \scriptsize{($\downarrow$)}&	Consistency \scriptsize{($\uparrow$)} &SR \scriptsize{($\uparrow$)}&	$\Delta$G \scriptsize{($\downarrow$)}&	Consistency \scriptsize{($\uparrow$)}&SR \scriptsize{($\uparrow$)}&	$\Delta$G \scriptsize{($\downarrow$)}&	Consistency \scriptsize{($\uparrow$)}\\
\midrule
RFD+IF&36.84\%&	-42.04&	63.15\%&	56.52\%&	-35.35&	65.21\%&	3.84\%&	-59.37&	53.84\%\\
APM&29.82\%&	-37.02&	54.38\%&43.47\%&	-31.04&	82.61\%&	0.0	&-61.54&	61.53\%\\
Complexa&54.38\%&	-45.81&	73.68\%	&78.26\%&	-41.32&	86.95\%&	\textbf{19.23\%}&	-66.56&	\textbf{88.46\%}\\
\model &\textbf{61.40\%}&	\textbf{-50.87}&	\textbf{98.24\%}&	\textbf{86.95\%}&	\textbf{-46.92}&	\textbf{100\%}&7.69\%&	\textbf{-68.82}&	30.76\%\\
\midrule
\multirow{2}{*}{Models} & \multicolumn{3}{c}{TGFb (Seen) } & \multicolumn{3}{c}{VirB8 (Seen)} & \multicolumn{3}{c}{EGFR (Zero-shot)}  \\
\cmidrule(r){2-4} \cmidrule(r){5-7} \cmidrule(r){8-10}
& SR \scriptsize{($\uparrow$)}&	$\Delta$G \scriptsize{($\downarrow$)}&	Consistency \scriptsize{($\uparrow$)} &SR \scriptsize{($\uparrow$)}&	$\Delta$G \scriptsize{($\downarrow$)}&	Consistency \scriptsize{($\uparrow$)}&SR \scriptsize{($\uparrow$)}&	$\Delta$G \scriptsize{($\downarrow$)}&	Consistency \scriptsize{($\uparrow$)}\\
\midrule
RFD+IF&11.11\%&	-31.09&	\textbf{55.56\%}&	14.28\%&	-32.91&	57.14\%&	\textbf{25.00\%}&	-46.87&	25.00\%\\
APM&11.11\%&	-23.16&11.11\%&	0.0	&-19.59&	42.85\%	&0.0&	-32.76&	50.00\%\\
Complexa&22.22\%	&-46.54&	33.33\%&14.28\%&	-38.56&	85.71\%	&0.0&	-40.90&	50.00\%\\
\model &\textbf{22.22\%}&	\textbf{-49.29}&	22.22\%&	\textbf{14.28\%}&	\textbf{-48.31}&	\textbf{100\%}&	0.0	&\textbf{-48.35}&\textbf{100.0\%}\\
\midrule
\multirow{2}{*}{Models} & \multicolumn{3}{c}{H3 (Zero-shot) } & \multicolumn{3}{c}{IL7Ra (Zero-shot)} & \multicolumn{3}{c}{Tie2 (Zero-shot)}  \\
\cmidrule(r){2-4} \cmidrule(r){5-7} \cmidrule(r){8-10}
& SR \scriptsize{($\uparrow$)}&	$\Delta$G \scriptsize{($\downarrow$)}&	Consistency \scriptsize{($\uparrow$)} &SR \scriptsize{($\uparrow$)}&	$\Delta$G \scriptsize{($\downarrow$)}&	Consistency \scriptsize{($\uparrow$)}&SR \scriptsize{($\uparrow$)}&	$\Delta$G \scriptsize{($\downarrow$)}&	Consistency \scriptsize{($\uparrow$)}\\
\midrule
RFD+IF&52.63\%&	-35.93&	26.31\%	&14.28\%&	-40.08&	42.85\%&	\textbf{50.00\%} &	-38.95&	75.00\%\\
APM&47.38\%&	-36.45&	42.10\%&0.0	&-23.93&	57.14\%	&0.0&	-27.85&	50.00\%\\
Complexa&60.52\%&	-40.65&	\textbf{47.36\%}&	28.57\%&	-42.26&	71.42\%&	0.0&	-33.83&	50.00\%\\
\model &\textbf{97.36\%}&	\textbf{-47.08}&	31.57\%	&\textbf{42.85\%}&\textbf{-50.42}&	\textbf{100\%}&	0.0&	\textbf{-42.39}&	\textbf{100.00\%}\\
\midrule
\multirow{2}{*}{Models} & \multicolumn{3}{c}{TrkA (Zero-shot)} & \multicolumn{3}{c}{Average } & &\multirow{2}{*}{Novelty \scriptsize{($\uparrow$)}} & \multirow{2}{*}{Diversity \scriptsize{($\uparrow$)}} \\
\cmidrule(r){2-4} \cmidrule(r){5-7} 
& SR \scriptsize{($\uparrow$)}&	$\Delta$G \scriptsize{($\downarrow$)}&	Consistency \scriptsize{($\uparrow$)} &SR \scriptsize{($\uparrow$)}&	$\Delta$G \scriptsize{($\downarrow$)}&	Consistency \scriptsize{($\uparrow$)}&&\\
\midrule
RFD+IF&25.00\%&	-33.28&	50.00\%&	28.95\%&	-39.59&	51.41\%&&	70.69\%&	30.38\%\\
APM&25.00\%&	-31.92&	25.00\%	&15.67\%&	-32.53&	47.67\%&&	51.47\%&	29.02\%\\
Complexa&50.00\%&	-41.50&	50.00\%&	32.74\%	&-43.79&	63.69\%&&	78.69\%&	\textbf{42.69\%}\\
\model &\textbf{75.00\%}&	\textbf{-46.04}&	\textbf{75.00\%}&	\textbf{40.77\%}&	\textbf{-49.85}&	\textbf{75.78\%}&	&\textbf{81.24\%}	&29.08\%\\
\bottomrule
\end{tabular}}
\end{center}
\caption{Model performance on target-protein mini-binder complex design task. 
Consistency denotes sequence-structure co-designability.
Our \model achieves the best average sequence design success rate, structure binding energy and sequence-structure co-designability.}
\label{Table:mini_binder_complex_design_results}
\end{table*}

\subsection{General Protein-Protein Complex Design}
\label{Sec:generation complex design}
This task evaluates the foundational capability of \model to design protein-protein complexes across diverse target scaffolds. We train and evaluate \model on our curated \dataset. 
Top-1 candidate performance for all methods is summarized in Table~\ref{Table:general_protein_protein_complex_design_performance}.
\textbf{\model demonstrates the best performance in generating high-affinity, structurally consistent binders across a wide spectrum of targets.} Specifically, \model achieves top performance across all metrics spanning sequence quality, structural reliability, and sequence-structure co-designability, reaching a sequence design success rate of 47.83\%, an interface binding energy of $-67.8$, and a co-designability score of 65.22\%. Although sequence diversity and novelty scores are slightly lower than those of Proteína-Complexa, \model achieves higher structural diversity (Appendix Table~\ref{Tab: strucrture_design_diversity_on_general_complex}). This indicates that \model successfully generates binders with flexible loop configurations and distinct subdomain topologies, whereas binders from Proteína-Complexa have almost invariant secondary structure cores. 

\subsection{Target Protein-Mini Binder Complex Design}
\label{Sec:mini_binder_design}

\noindent\textbf{Datasets.} We curate experimentally validated target-protein mini-binder complexes across ten diverse target classes from \citet{bennett2023improving}. For targets with more than 50 complexes, we partition the data using a random 8:1:1 train/validation/test split. For targets with fewer complexes, all instances are reserved for testing to evaluate model generalization in a zero-shot setting. Comprehensive dataset statistics are provided in Appendix Table~\ref{Table:target-protein mini-binder dataset statistics}.

\noindent \textbf{Main Results.} 
The results for target-protein mini-binder complex design are summarized in Table~\ref{Table:mini_binder_complex_design_results}, with target-specific breakdowns provided in Appendix Tables~\ref{Table:FGFR2}\text{--}\ref{Table:TrkA}. \textbf{Overall, \model demonstrates superior capacity for generating novel, high-affinity binders across diverse target classes.} Specifically, \model achieves higher sequence design success rates on 7 out of 10 targets, superior structure binding energy across all 10 targets, and improved co-designability on 7 out of 10 targets. Notably, \model also yields higher novelty scores, reversing the trend observed in our general complex design task. We attribute this shift to the distinctions between the fine-tuning dataset and \dataset, enabling \model to navigate a broader candidate space for novel binder discovery.

\begin{table*}[!t]
\scriptsize
\begin{center}
\setlength{\tabcolsep}{0.46mm}{
\begin{tabular}{lccccccccc}
\toprule
\multirow{2}{*}{Models} & \multicolumn{6}{c}{Sequence } & \multicolumn{2}{c}{Structure } & \multirow{2}{*}{Co-designability \scriptsize{($\uparrow$)}}  \\
\cmidrule(r){2-7} \cmidrule(r){8-9} 
& ipTM \scriptsize{($\uparrow$)}	& pLDDT \scriptsize{($\uparrow$)}	& SR \scriptsize{($\uparrow$)}	& H1-Novelty \scriptsize{($\uparrow$)} &	H2-Novelty \scriptsize{($\uparrow$)} &	H3-Novelty \scriptsize{($\uparrow$)} &	$\Delta$G \scriptsize{($\downarrow$)}	& Designability \scriptsize{($\uparrow$)} &  \\
\midrule
RFD+IF&0.508&	80.05&	16.79\%&	47.30\%&	56.22\%&	61.98\%&	-33.09&	59.29\%&	59.29\%\\
APM&0.256&	77.92&	6.61\%&	45.71\%&	54.17\%&	64.43\%&	-26.68&	53.57\%&	52.74\%\\
Complexa&0.515&	81.08&	16.92\%&	\textbf{53.80\%}&	66.26\%&	71.91\%&	-32.80&	90.26\%&	92.32\%\\
\model &\textbf{0.526}&	\textbf{81.41}&	\textbf{18.77\%}&	52.31\%&	\textbf{67.85\%}&	\textbf{71.93\%}&	\textbf{-44.46}&	\textbf{98.34\%}&	\textbf{99.39\%}\\
\bottomrule
\end{tabular}}
\end{center}
\caption{Model performance on antigen-antibody complex design task. 
Our \model achieves the best sequence design success rate, structure binding energy and sequence-structure co-designability.}
\label{Table:antigen_antibody_complex_design_performance}
\end{table*}

\subsection{Antigen-Antibody Complex Design}
\label{Sec:antibody_design}

\textbf{Dataset.}
Following \citet{jiniterative}, we curate $4{,}261$ complete antigen-antibody complexes from the Structural Antibody Database~(\citet{dunbar2014sabdab}, SAbDab) by removing cases lacking a light chain or antigen. Train/validation/test splits (8:1:1 ratio) are constructed by clustering individual CDR sequences at 40\% identity using MMseqs2 \citep{steinegger2017mmseqs2}, yielding 1,034 (CDR-H1), 1,418 (CDR-H2), and 2,254 (CDR-H3) clusters. Full split statistics are provided in Appendix Table~\ref{Tab: data_statistics_antibody_antigen_complex}.

\noindent\textbf{Main Results.}
Average performance across distinct CDR cluster splits for antigen-antibody complex design is summarized in Table~\ref{Table:antigen_antibody_complex_design_performance}, with CDR-specific breakdowns detailed in Appendix Tables~\ref{Table:antigen_antibody_complex_design_performance_cdrh1}\text{--}\ref{Table:antigen_antibody_complex_design_performance_cdrh3}. \textbf{Overall, \model outperforms all baselines in generating high-affinity and novel antibodies.} Specifically, \model achieves top performance across sequence design success rate, CDR H2/H3 novelty, interface binding energy ($\Delta G$), and sequence-structure co-designability. Crucially, both the designability and co-designability of \model exceed $95\%$, demonstrating exceptional structure fidelity and tight sequence-structure consistency.

\begin{figure*}
\begin{minipage}[t]{0.25\linewidth}
\centering
\includegraphics[width=4.5cm]{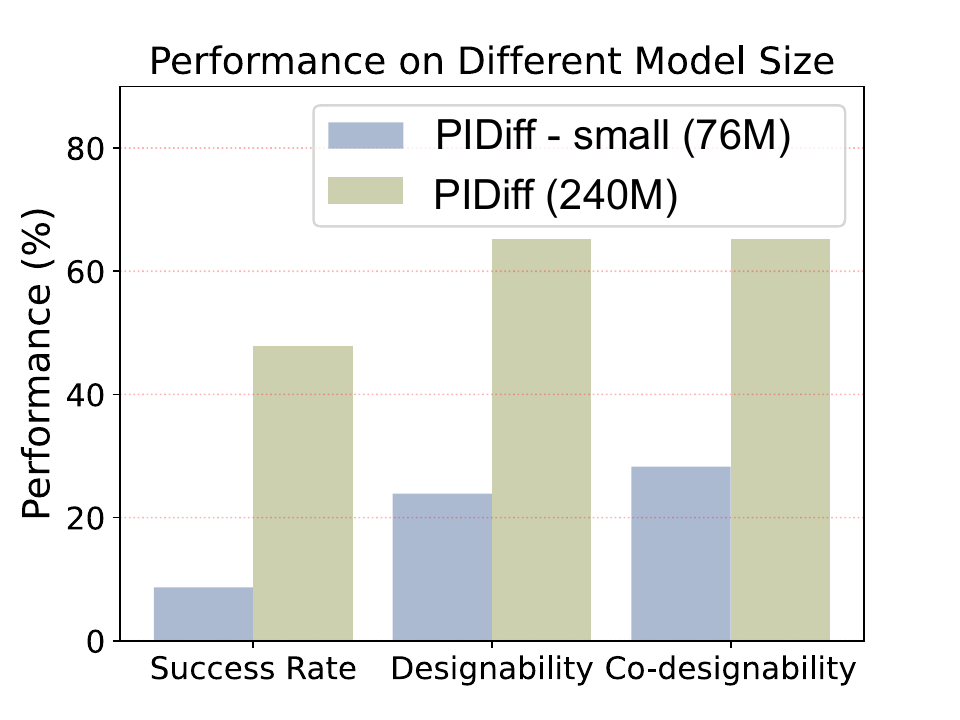}
\centerline{\small{(a) Model Scaling}}
\end{minipage}%
\begin{minipage}[t]{0.25\linewidth}
\centering
\includegraphics[width=4.5cm]{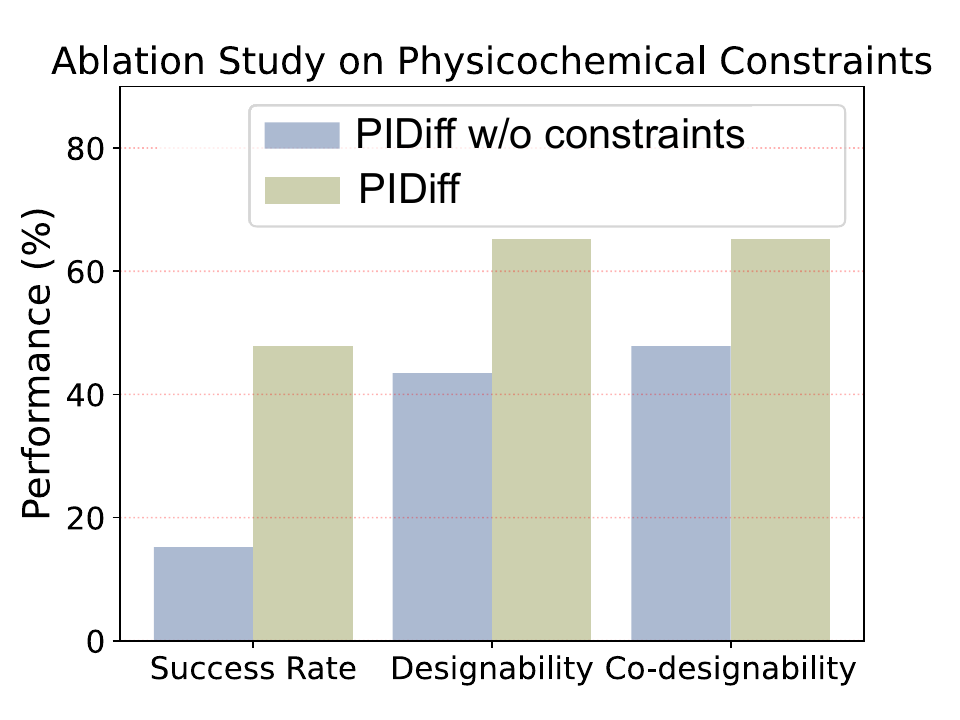}
\centerline{\small{(b) Physicochemical Properties}}
\end{minipage}%
\begin{minipage}[t]{0.25\linewidth}
\centering
\includegraphics[width=4.5cm]{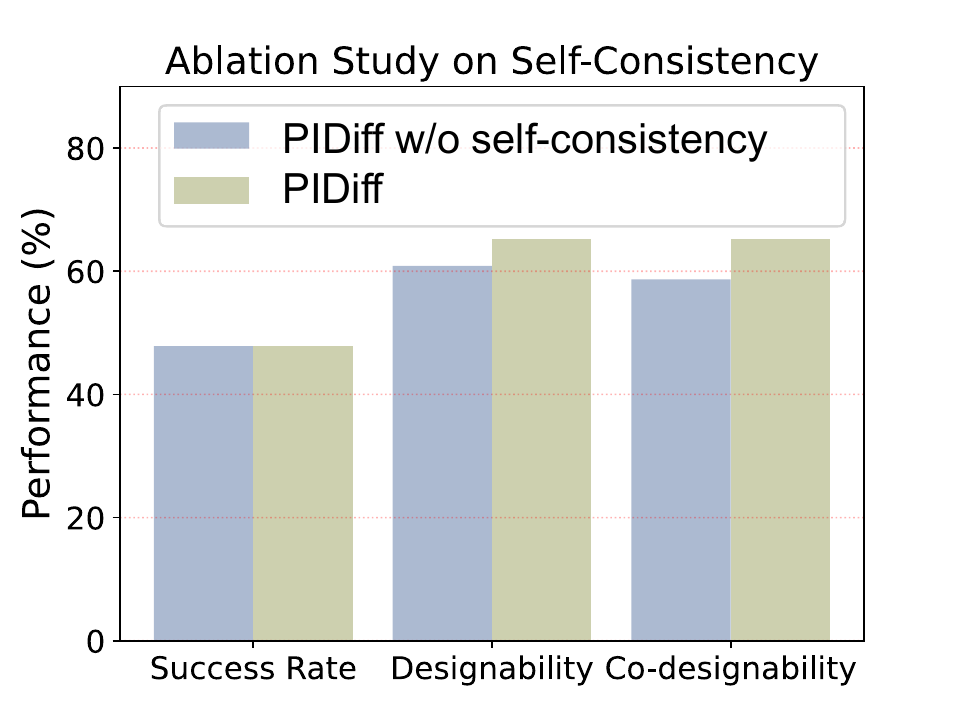}
\centerline{\small{(c) Self-Consistency}}
\end{minipage}%
\begin{minipage}[t]{0.25\linewidth}
\centering
\includegraphics[width=4.5cm]{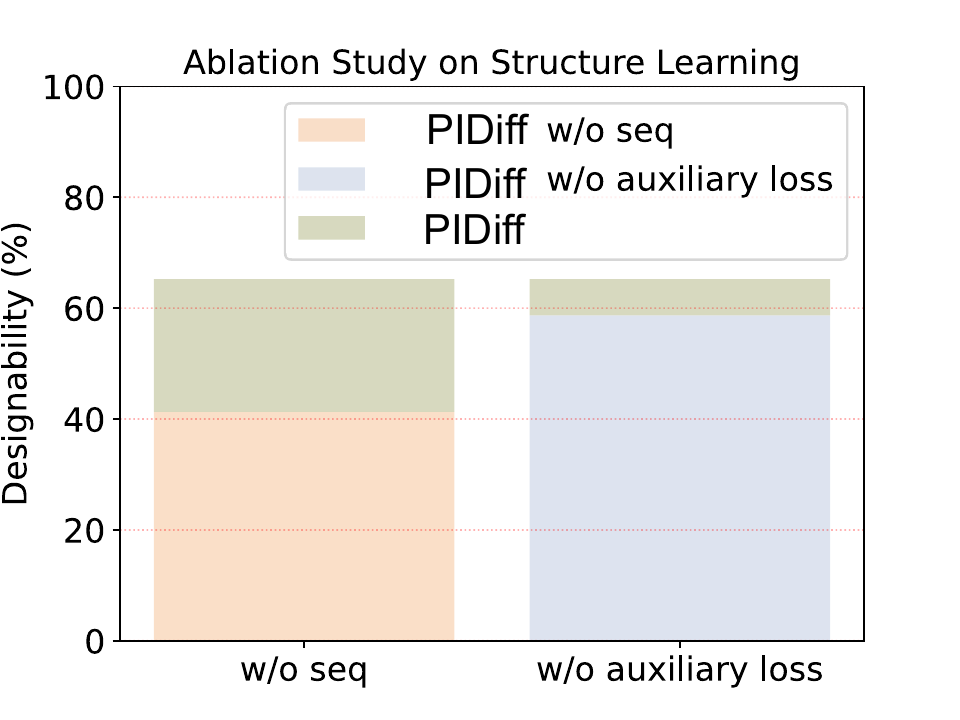}
\centerline{\small{(d) Structure Learning}}
\end{minipage}
\begin{minipage}[t]{0.27\linewidth}
\centering
\includegraphics[width=4.8cm]{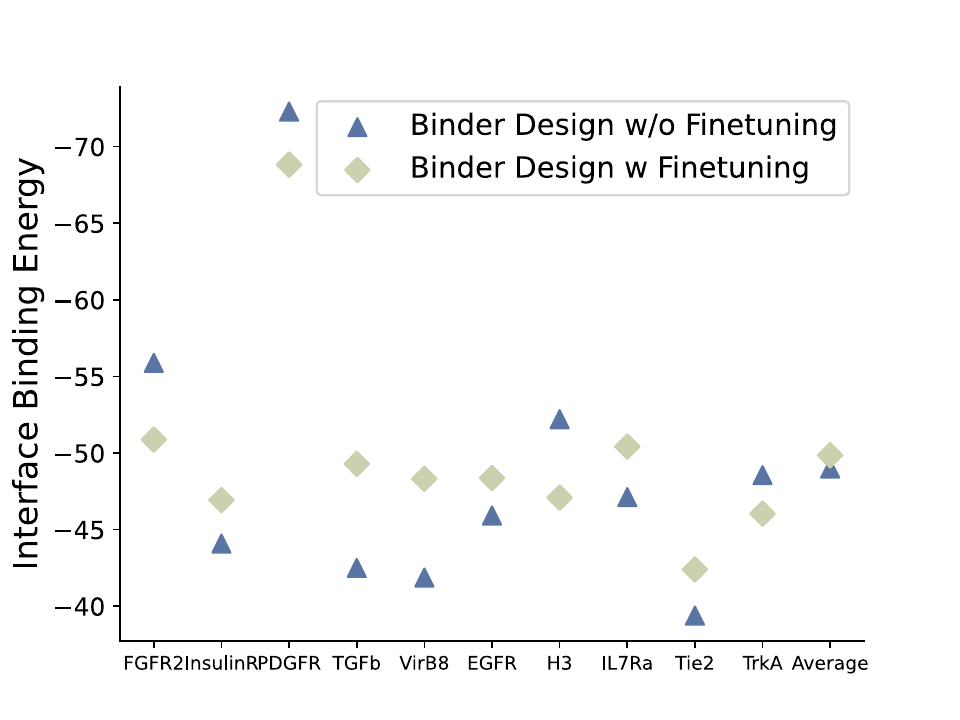}
\centerline{\small{(e) Binder Finetuning}}
\end{minipage}%
\begin{minipage}[t]{0.24\linewidth}
\centering
\includegraphics[width=3.8cm]{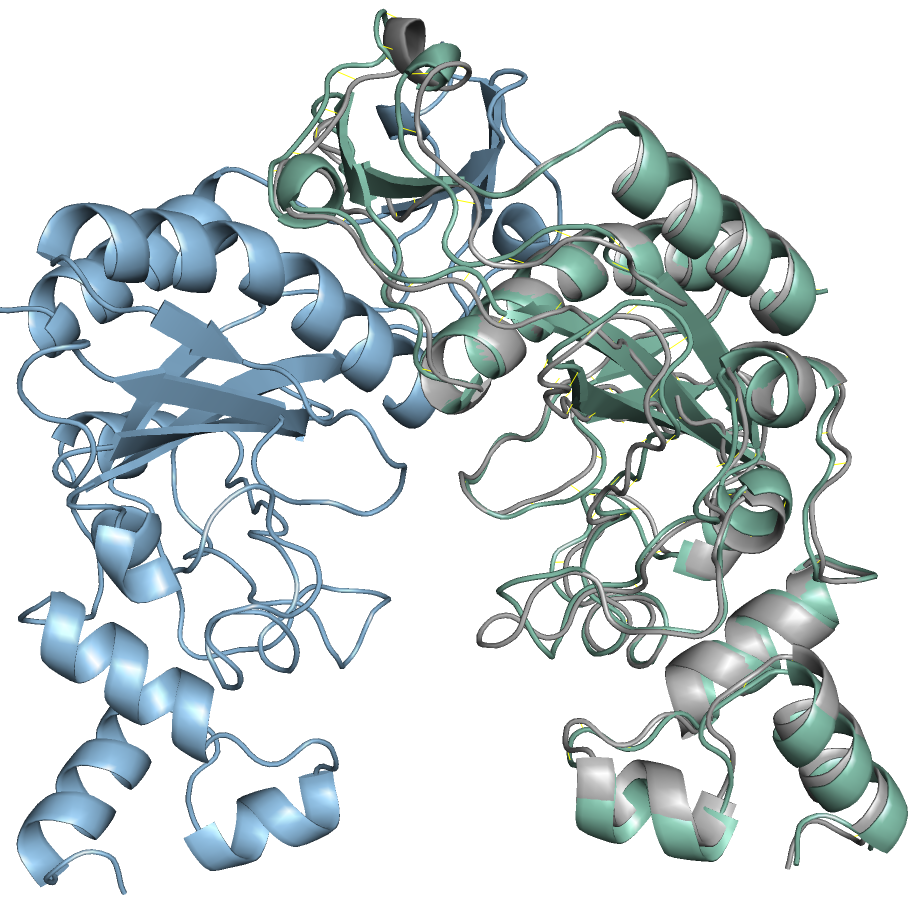}
\centerline{\small{(f) General Protein Complex}}
\end{minipage}%
\begin{minipage}[t]{0.24\linewidth}
\centering
\includegraphics[width=3.8cm]{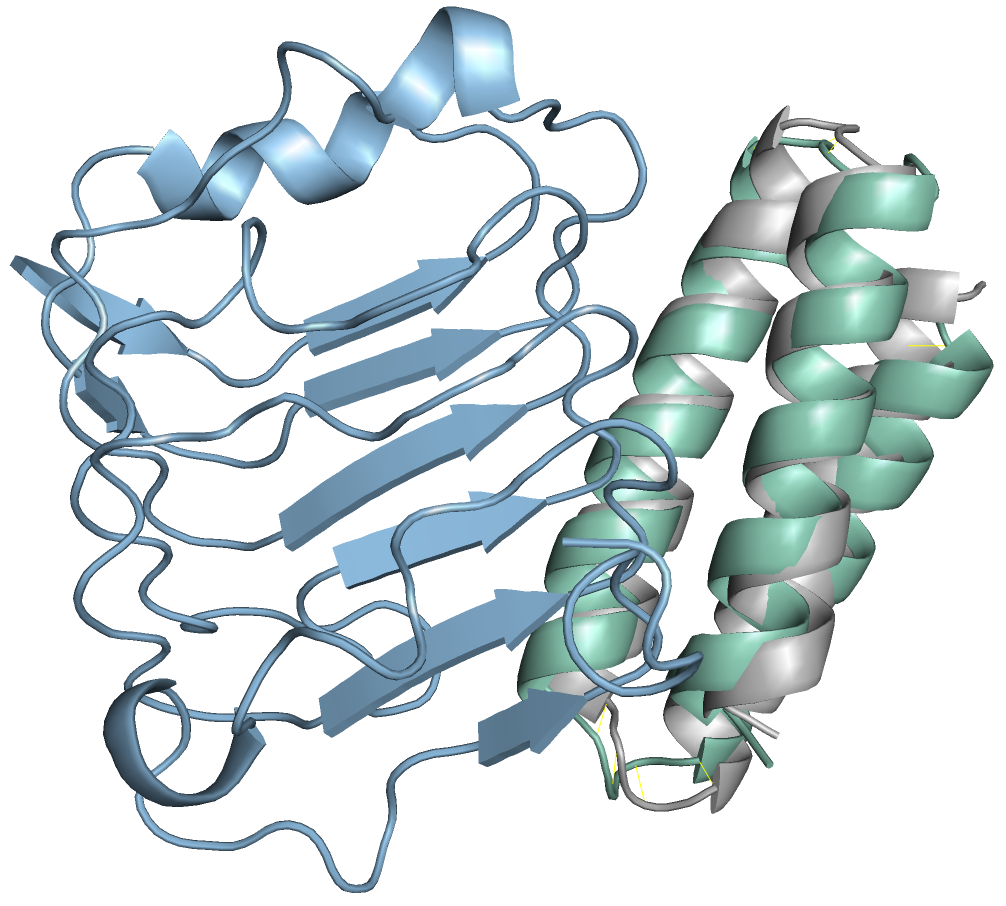}
\centerline{\small{(g) Binder Design}}
\end{minipage}%
\begin{minipage}[t]{0.25\linewidth}
\centering
\includegraphics[width=3.8cm]{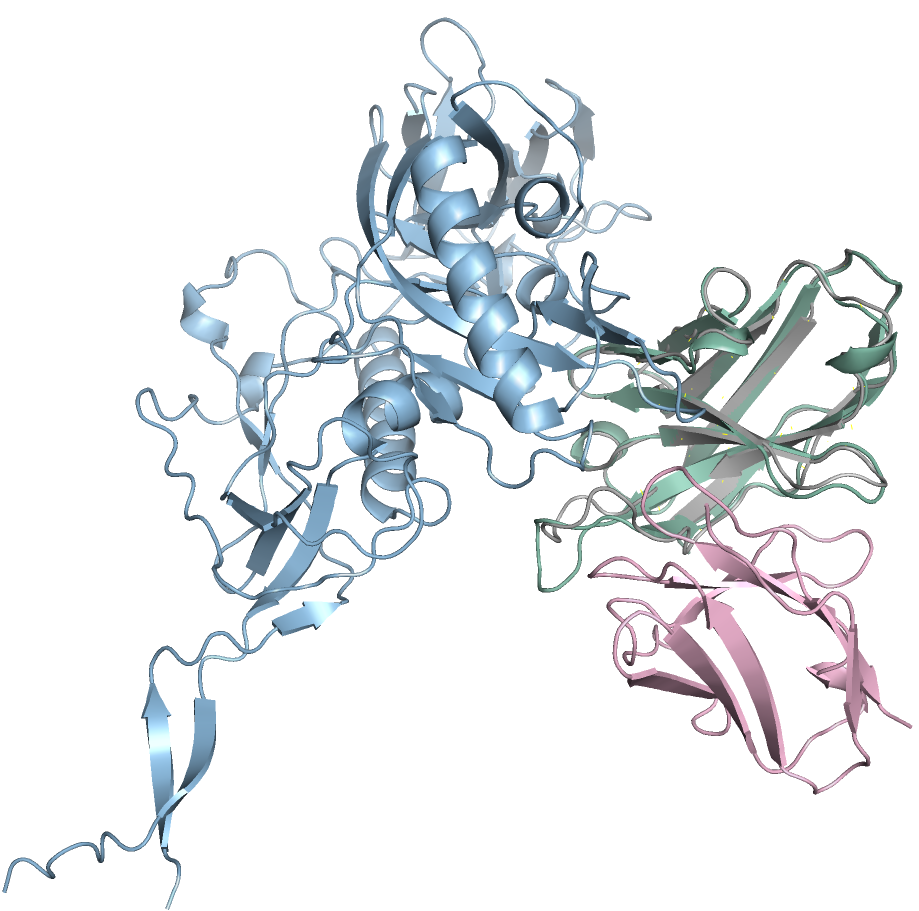}
\centerline{\small{(h) Antibody Design}}
\end{minipage}
	\caption{
    Ablation study: (a) - (e), and case study with target protein (f) PDB 1STX.E, (g) InsulinR, (h) HIV-1 (7UCG). Target proteins are shown in blue, the folded structures for the designed binder sequences in green, and the designed binder structures in gray, with light chain in pink for antibody.
    } 
 \label{Fig:ablation_study}
\end{figure*}

\section{Analysis}
\subsection{Does Model Scale Help?}
\textbf{\model scales effectively with parameter capacity.} We benchmark our 240M \model against a smaller model (76M) configured with 4 blocks and a hidden dimension of 480. As shown in Figure~\ref{Fig:ablation_study}(a), scaling capacity improves sequence success rate, structural designability, and co-designability on \dataset. These results highlight the scalability of our architecture, indicating that performance scales predictably with model capacity.

\subsection{Effect of Physicochemical Complementarity Constraints}
\label{Sec:effect_of_physicochemical_constraints}
As stable protein-protein interactions depend on complementary geometries and physicochemical properties, we conduct an ablation study by training a variant of \model without physicochemical features on \dataset. As illustrated in Figure~\ref{Fig:ablation_study}(b), omitting these constraints leads to a significant performance degradation across both sequence and structural metrics. \textbf{These findings confirm that explicitly incorporating  physicochemical features is critical for stable interface design.}

\subsection{Influence of Self-Consistency}
\label{Sec:influence_of_self_consistency}
To evaluate the role of self-consistency during inference, we conduct an ablation by disabling this mechanism on the general protein-protein complex design task. As shown in Figure~\ref{Fig:ablation_study}(c), removing self-consistency yields a measurable drop in both structural designability and co-designability. \textbf{This validates self-consistency as an effective inference-time strategy for generating reliable structures and reinforcing sequence-structure co-design alignment.}

\subsection{Impact of Joint Co-Design and Auxiliary Losses}
\label{Sec:impact_of_codesign_and_auxiliary_loss}
To isolate the contributions of the joint sequence-structure co-design mechanism and auxiliary supervision, we evaluate a structure-only ablation (\model\ w/o seq) alongside a model variant trained without auxiliary loss terms (\model\ w/o auxiliary loss). As illustrated in Figure~\ref{Fig:ablation_study}(d), both components enhance structural designability, with sequence-structure co-design exerting a significantly stronger effect. We attribute these gains to the stricter physical regularization provided by both mechanisms, \textbf{confirming that they are vital for designing reliable structures.}

\subsection{Impact of Supervised Fine-Tuning}
\label{Sec:impact_of_finetuning}
To evaluate how domain-specific adaptation affects downstream performance, we compare our pretrained \model on the mini-binder design task. As shown in Figure~\ref{Fig:ablation_study}(e), targeted fine-tuning improves interface binding energy ($\Delta G$) across 6 out of 10 targets, slightly promoting the mean energy score from $-48.98$ to $-49.85$. Full results are provided in Appendix Table~\ref{Table:mini_binder_complex_design_results_on_pretrain}. \textbf{These results demonstrate that while our pretrained model already possesses robust zero-shot design capabilities, supervised fine-tuning further enhances binding affinity for specialized applications.}

\subsection{Designing Diverse and High-Affinity Protein Complexes}
\label{Sec:qualitative_case_studies}
Figures~\ref{Fig:ablation_study}(f)\text{--}(h) showcase qualitative case studies across general complex, mini-binder, and antibody design tasks. Spanning distinct target topologies, the generated candidates consistently satisfy stringent quality thresholds: ipTM $> 0.8$, pLDDT $> 85$, interface binding energy $< -35$, and co-design RMSD $< 2\,\text{\AA}$. \textbf{This highlights \model's ability to co-design highly consistent binder sequences and structures against diverse target scaffolds.} 

\section{Conclusion}
In this work, we introduce \model, a physicochemically enhanced multimodal diffusion method for joint binder protein sequence-structure co-design. Powered by \network, an architecture combining IPA layer with sequence-level self-attention, \model effectively couples 3D  geometry with amino acid dependencies. Across general protein complex generation as well as specialized mini-binder and antibody design tasks, \model consistently surpasses state-of-the-art baselines across multiple metrics. Overall, \model demonstrates strong potential as a robust and scalable tool for computational binder discovery. A potential limitation is the absence of wet-lab validation, which is an essential step planned for our future research to further validate the model's practical utility.


\section*{Acknowledgments}
We thank Lambda, Inc. for generously providing the computational resources that supported this work.

\bibliography{icml2025}

@article{jumper2021highly,
  title={Highly accurate protein structure prediction with AlphaFold},
  author={Jumper, John and Evans, Richard and Pritzel, Alexander and Green, Tim and Figurnov, Michael and Ronneberger, Olaf and Tunyasuvunakool, Kathryn and Bates, Russ and {\v{Z}}{\'\i}dek, Augustin and Potapenko, Anna and others},
  journal={nature},
  volume={596},
  number={7873},
  pages={583--589},
  year={2021},
  publisher={Nature Publishing Group UK London}
}

@article{ho2020denoising,
  title={Denoising diffusion probabilistic models},
  author={Ho, Jonathan and Jain, Ajay and Abbeel, Pieter},
  journal={Advances in neural information processing systems},
  volume={33},
  pages={6840--6851},
  year={2020}
}

@article{austin2021structured,
  title={Structured denoising diffusion models in discrete state-spaces},
  author={Austin, Jacob and Johnson, Daniel D and Ho, Jonathan and Tarlow, Daniel and Van Den Berg, Rianne},
  journal={Advances in neural information processing systems},
  volume={34},
  pages={17981--17993},
  year={2021}
}

@article{de2022riemannian,
  title={Riemannian score-based generative modelling},
  author={De Bortoli, Valentin and Mathieu, Emile and Hutchinson, Michael and Thornton, James and Teh, Yee Whye and Doucet, Arnaud},
  journal={Advances in neural information processing systems},
  volume={35},
  pages={2406--2422},
  year={2022}
}

@article{yim2023se,
  title={SE (3) diffusion model with application to protein backbone generation},
  author={Yim, Jason and Trippe, Brian L and De Bortoli, Valentin and Mathieu, Emile and Doucet, Arnaud and Barzilay, Regina and Jaakkola, Tommi},
  journal={arXiv preprint arXiv:2302.02277},
  year={2023}
}

@book{hsu2002stochastic,
  title={Stochastic analysis on manifolds},
  author={Hsu, Elton P},
  number={38},
  year={2002},
  publisher={American Mathematical Soc.}
}

@article{song2020score,
  title={Score-based generative modeling through stochastic differential equations},
  author={Song, Yang and Sohl-Dickstein, Jascha and Kingma, Diederik P and Kumar, Abhishek and Ermon, Stefano and Poole, Ben},
  journal={arXiv preprint arXiv:2011.13456},
  year={2020}
}

@article{nelson2010development,
  title={Development trends for human monoclonal antibody therapeutics},
  author={Nelson, Aaron L and Dhimolea, Eugen and Reichert, Janice M},
  journal={Nature reviews drug discovery},
  volume={9},
  number={10},
  pages={767--774},
  year={2010},
  publisher={Nature Publishing Group UK London}
}

@article{brennan2010antibody,
  title={Antibody-based proteomics: fast-tracking molecular diagnostics in oncology},
  author={Brennan, Donal J and O'connor, Darran P and Rexhepaj, Elton and Ponten, Fredrik and Gallagher, William M},
  journal={Nature Reviews Cancer},
  volume={10},
  number={9},
  pages={605--617},
  year={2010},
  publisher={Nature Publishing Group UK London}
}

@article{correia2014proof,
  title={Proof of principle for epitope-focused vaccine design},
  author={Correia, Bruno E and Bates, John T and Loomis, Rebecca J and Baneyx, Gretchen and Carrico, Chris and Jardine, Joseph G and Rupert, Peter and Correnti, Colin and Kalyuzhniy, Oleksandr and Vittal, Vinayak and others},
  journal={Nature},
  volume={507},
  number={7491},
  pages={201--206},
  year={2014},
  publisher={Nature Publishing Group UK London}
}

@article{sesterhenn2020novo,
  title={De novo protein design enables the precise induction of RSV-neutralizing antibodies},
  author={Sesterhenn, Fabian and Yang, Che and Bonet, Jaume and Cramer, Johannes T and Wen, Xiaolin and Wang, Yimeng and Chiang, Chi-I and Abriata, Luciano A and Kucharska, Iga and Castoro, Giacomo and others},
  journal={Science},
  volume={368},
  number={6492},
  pages={eaay5051},
  year={2020},
  publisher={American Association for the Advancement of Science}
}

@article{chothia1975principles,
  title={Principles of protein--protein recognition},
  author={Chothia, Cyrus and Janin, Jo{\"e}l},
  journal={Nature},
  volume={256},
  number={5520},
  pages={705--708},
  year={1975},
  publisher={Nature Publishing Group UK London}
}

@article{conte1999atomic,
  title={The atomic structure of protein-protein recognition sites},
  author={Conte, Loredana Lo and Chothia, Cyrus and Janin, Jo{\"e}l},
  journal={Journal of molecular biology},
  volume={285},
  number={5},
  pages={2177--2198},
  year={1999},
  publisher={Elsevier}
}

@article{gainza2023novo,
  title={De novo design of protein interactions with learned surface fingerprints},
  author={Gainza, Pablo and Wehrle, Sarah and Van Hall-Beauvais, Alexandra and Marchand, Anthony and Scheck, Andreas and Harteveld, Zander and Buckley, Stephen and Ni, Dongchun and Tan, Shuguang and Sverrisson, Freyr and others},
  journal={Nature},
  volume={617},
  number={7959},
  pages={176--184},
  year={2023},
  publisher={Nature Publishing Group UK London}
}

@article{chevalier2017massively,
  title={Massively parallel de novo protein design for targeted therapeutics},
  author={Chevalier, Aaron and Silva, Daniel-Adriano and Rocklin, Gabriel J and Hicks, Derrick R and Vergara, Renan and Murapa, Patience and Bernard, Steffen M and Zhang, Lu and Lam, Kwok-Ho and Yao, Guorui and others},
  journal={Nature},
  volume={550},
  number={7674},
  pages={74--79},
  year={2017},
  publisher={Nature Publishing Group UK London}
}

@article{silva2019novo,
  title={De novo design of potent and selective mimics of IL-2 and IL-15},
  author={Silva, Daniel-Adriano and Yu, Shawn and Ulge, Umut Y and Spangler, Jamie B and Jude, Kevin M and Lab{\~a}o-Almeida, Carlos and Ali, Lestat R and Quijano-Rubio, Alfredo and Ruterbusch, Mikel and Leung, Isabel and others},
  journal={Nature},
  volume={565},
  number={7738},
  pages={186--191},
  year={2019},
  publisher={Nature Publishing Group UK London}
}

@article{cao2022design,
  title={Design of protein-binding proteins from the target structure alone},
  author={Cao, Longxing and Coventry, Brian and Goreshnik, Inna and Huang, Buwei and Sheffler, William and Park, Joon Sung and Jude, Kevin M and Markovi{\'c}, Iva and Kadam, Rameshwar U and Verschueren, Koen HG and others},
  journal={Nature},
  volume={605},
  number={7910},
  pages={551--560},
  year={2022},
  publisher={Nature Publishing Group UK London}
}

@article{evans2021protein,
  title={Protein complex prediction with AlphaFold-Multimer},
  author={Evans, Richard and O’Neill, Michael and Pritzel, Alexander and Antropova, Natasha and Senior, Andrew and Green, Tim and {\v{Z}}{\'\i}dek, Augustin and Bates, Russ and Blackwell, Sam and Yim, Jason and others},
  journal={biorxiv},
  pages={2021--10},
  year={2021},
  publisher={Cold Spring Harbor Laboratory}
}

@article{bennett2023improving,
  title={Improving de novo protein binder design with deep learning},
  author={Bennett, Nathaniel R and Coventry, Brian and Goreshnik, Inna and Huang, Buwei and Allen, Aza and Vafeados, Dionne and Peng, Ying Po and Dauparas, Justas and Baek, Minkyung and Stewart, Lance and others},
  journal={Nature Communications},
  volume={14},
  number={1},
  pages={2625},
  year={2023},
  publisher={Nature Publishing Group UK London}
}

@article{zambaldi2024novo,
  title={De novo design of high-affinity protein binders with AlphaProteo},
  author={Zambaldi, Vinicius and La, David and Chu, Alexander E and Patani, Harshnira and Danson, Amy E and Kwan, Tristan OC and Frerix, Thomas and Schneider, Rosalia G and Saxton, David and Thillaisundaram, Ashok and others},
  journal={arXiv preprint arXiv:2409.08022},
  year={2024}
}

@article{vazquez2024novo,
  title={De novo design of high-affinity binders of bioactive helical peptides},
  author={V{\'a}zquez Torres, Susana and Leung, Philip JY and Venkatesh, Preetham and Lutz, Isaac D and Hink, Fabian and Huynh, Huu-Hien and Becker, Jessica and Yeh, Andy Hsien-Wei and Juergens, David and Bennett, Nathaniel R and others},
  journal={Nature},
  volume={626},
  number={7998},
  pages={435--442},
  year={2024},
  publisher={Nature Publishing Group UK London}
}

@article{wu2024sequence,
  title={Sequence-specific targeting of intrinsically disordered protein regions},
  author={Wu, Kejia and Jiang, Hanlun and Hicks, Derrick R and Liu, Caixuan and Muratspahi{\'c}, Edin and Ramelot, Theresa A and Liu, Yuexuan and McNally, Kerrie and Gaur, Amit and Coventry, Brian and others},
  journal={bioRxiv},
  pages={2024--07},
  year={2024},
  publisher={Cold Spring Harbor Laboratory}
}

@article{vaswani2017attention,
  title={Attention is all you need},
  author={Vaswani, Ashish and Shazeer, Noam and Parmar, Niki and Uszkoreit, Jakob and Jones, Llion and Gomez, Aidan N and Kaiser, {\L}ukasz and Polosukhin, Illia},
  journal={Advances in neural information processing systems},
  volume={30},
  year={2017}
}

@article{watson2023novo,
  title={De novo design of protein structure and function with RFdiffusion},
  author={Watson, Joseph L and Juergens, David and Bennett, Nathaniel R and Trippe, Brian L and Yim, Jason and Eisenach, Helen E and Ahern, Woody and Borst, Andrew J and Ragotte, Robert J and Milles, Lukas F and others},
  journal={Nature},
  volume={620},
  number={7976},
  pages={1089--1100},
  year={2023},
  publisher={Nature Publishing Group UK London}
}

@article{dauparas2022robust,
  title={Robust deep learning--based protein sequence design using ProteinMPNN},
  author={Dauparas, Justas and Anishchenko, Ivan and Bennett, Nathaniel and Bai, Hua and Ragotte, Robert J and Milles, Lukas F and Wicky, Basile IM and Courbet, Alexis and de Haas, Rob J and Bethel, Neville and others},
  journal={Science},
  volume={378},
  number={6615},
  pages={49--56},
  year={2022},
  publisher={American Association for the Advancement of Science}
}

@article{chen2025all,
  title={An all-atom generative model for designing protein complexes},
  author={Chen, Ruizhe and Xue, Dongyu and Zhou, Xiangxin and Zheng, Zaixiang and Zeng, Xiangxiang and Gu, Quanquan},
  journal={arXiv preprint arXiv:2504.13075},
  year={2025}
}

@inproceedings{didi2026scaling,
  title={Scaling atomistic protein binder design with generative pretraining and test-time compute},
  author={Didi, Kieran and Zhang, Zuobai and Zhou, Guoqing and Reidenbach, Danny and Cao, Zhonglin and Cha, Sooyoung and Geffner, Tomas and Dallago, Christian and Tang, Jian and Bronstein, Michael and others},
  booktitle={International Conference on Learning Representations},
  volume={2026},
  pages={61056--61109},
  year={2026}
}

@article{vazquez2015challenges,
  title={Challenges and opportunities for non-antibody scaffold drugs},
  author={Vazquez-Lombardi, Rodrigo and Phan, Tri Giang and Zimmermann, Carsten and Lowe, David and Jermutus, Lutz and Christ, Daniel},
  journal={Drug discovery today},
  volume={20},
  number={10},
  pages={1271--1283},
  year={2015},
  publisher={Elsevier}
}

@article{chao2006isolating,
  title={Isolating and engineering human antibodies using yeast surface display},
  author={Chao, Ginger and Lau, Wai L and Hackel, Benjamin J and Sazinsky, Stephen L and Lippow, Shaun M and Wittrup, K Dane},
  journal={Nature protocols},
  volume={1},
  number={2},
  pages={755--768},
  year={2006},
  publisher={Nature Publishing Group}
}

@article{hackel2008picomolar,
  title={Picomolar affinity fibronectin domains engineered utilizing loop length diversity, recursive mutagenesis, and loop shuffling},
  author={Hackel, Benjamin J and Kapila, Atul and Wittrup, K Dane},
  journal={Journal of molecular biology},
  volume={381},
  number={5},
  pages={1238--1252},
  year={2008},
  publisher={Elsevier}
}

@article{fleishman2011computational,
  title={Computational design of proteins targeting the conserved stem region of influenza hemagglutinin},
  author={Fleishman, Sarel J and Whitehead, Timothy A and Ekiert, Damian C and Dreyfus, Cyrille and Corn, Jacob E and Strauch, Eva-Maria and Wilson, Ian A and Baker, David},
  journal={Science},
  volume={332},
  number={6031},
  pages={816--821},
  year={2011},
  publisher={American Association for the Advancement of Science}
}

@article{humphreys2021computed,
  title={Computed structures of core eukaryotic protein complexes},
  author={Humphreys, Ian R and Pei, Jimin and Baek, Minkyung and Krishnakumar, Aditya and Anishchenko, Ivan and Ovchinnikov, Sergey and Zhang, Jing and Ness, Travis J and Banjade, Sudeep and Bagde, Saket R and others},
  journal={Science},
  volume={374},
  number={6573},
  pages={eabm4805},
  year={2021},
  publisher={American Association for the Advancement of Science}
}

@article{bryant2022improved,
  title={Improved prediction of protein-protein interactions using AlphaFold2},
  author={Bryant, Patrick and Pozzati, Gabriele and Elofsson, Arne},
  journal={Nature communications},
  volume={13},
  number={1},
  pages={1265},
  year={2022},
  publisher={Nature Publishing Group UK London}
}

@article{pacesa2024bindcraft,
  title={BindCraft: one-shot design of functional protein binders},
  author={Pacesa, Martin and Nickel, Lennart and Schellhaas, Christian and Schmidt, Joseph and Pyatova, Ekaterina and Kissling, Lucas and Barendse, Patrick and Choudhury, Jagrity and Kapoor, Srajan and Alcaraz-Serna, Ana and others},
  journal={bioRxiv},
  pages={2024--09},
  year={2024},
  publisher={Cold Spring Harbor Laboratory}
}

@article{berger2024preclinical,
  title={Preclinical proof of principle for orally delivered Th17 antagonist miniproteins},
  author={Berger, Stephanie and Seeger, Franziska and Yu, Ta-Yi and Aydin, Merve and Yang, Huilin and Rosenblum, Daniel and Guenin-Mac{\'e}, Laure and Glassman, Caleb and Arguinchona, Lauren and Sniezek, Catherine and others},
  journal={Cell},
  volume={187},
  number={16},
  pages={4305--4317},
  year={2024},
  publisher={Elsevier}
}

@inproceedings{sohl2015deep,
  title={Deep unsupervised learning using nonequilibrium thermodynamics},
  author={Sohl-Dickstein, Jascha and Weiss, Eric and Maheswaranathan, Niru and Ganguli, Surya},
  booktitle={International conference on machine learning},
  pages={2256--2265},
  year={2015},
  organization={PMLR}
}

@article{song2019generative,
  title={Generative modeling by estimating gradients of the data distribution},
  author={Song, Yang and Ermon, Stefano},
  journal={Advances in neural information processing systems},
  volume={32},
  year={2019}
}

@article{trippe2022diffusion,
  title={Diffusion probabilistic modeling of protein backbones in 3d for the motif-scaffolding problem},
  author={Trippe, Brian L and Yim, Jason and Tischer, Doug and Baker, David and Broderick, Tamara and Barzilay, Regina and Jaakkola, Tommi},
  journal={arXiv preprint arXiv:2206.04119},
  year={2022}
}

@article{wang2024diffusion,
  title={Diffusion language models are versatile protein learners},
  author={Wang, Xinyou and Zheng, Zaixiang and Ye, Fei and Xue, Dongyu and Huang, Shujian and Gu, Quanquan},
  journal={arXiv preprint arXiv:2402.18567},
  year={2024}
}

@article{alamdari2023protein,
  title={Protein generation with evolutionary diffusion: sequence is all you need},
  author={Alamdari, Sarah and Thakkar, Nitya and Van Den Berg, Rianne and Tenenholtz, Neil and Strome, Robert and Moses, Alan M and Lu, Alex X and Fusi, Nicol{\`o} and Amini, Ava P and Yang, Kevin K},
  journal={BioRxiv},
  pages={2023--09},
  year={2023},
  publisher={Cold Spring Harbor Laboratory}
}

@inproceedings{wang2025dplm,
  title={Dplm-2: A multimodal diffusion protein language model},
  author={Wang, Xinyou and Zheng, Zaixiang and Ye, Fei and Xue, Dongyu and Huang, Shujian and Gu, Quanquan},
  booktitle={International Conference on Learning Representations},
  volume={2025},
  pages={35463--35490},
  year={2025}
}

@article{song2025ppdiff,
  title={Ppdiff: Diffusing in hybrid sequence-structure space for protein-protein complex design},
  author={Song, Zhenqiao and Li, Tiaoxiao and Li, Lei and Min, Martin Renqiang},
  journal={arXiv preprint arXiv:2506.11420},
  year={2025}
}

@article{abramson2024accurate,
  title={Accurate structure prediction of biomolecular interactions with AlphaFold 3},
  author={Abramson, Josh and Adler, Jonas and Dunger, Jack and Evans, Richard and Green, Tim and Pritzel, Alexander and Ronneberger, Olaf and Willmore, Lindsay and Ballard, Andrew J and Bambrick, Joshua and others},
  journal={Nature},
  pages={1--3},
  year={2024},
  publisher={Nature Publishing Group UK London}
}

@article{lin2022language,
  title={Language models of protein sequences at the scale of evolution enable accurate structure prediction},
  author={Lin, Zeming and Akin, Halil and Rao, Roshan and Hie, Brian and Zhu, Zhongkai and Lu, Wenting and Smetanin, Nikita and dos Santos Costa, Allan and Fazel-Zarandi, Maryam and Sercu, Tom and Candido, Sal and others},
  journal={bioRxiv},
  year={2022},
  publisher={Cold Spring Harbor Laboratory}
}

@article{chaudhury2010pyrosetta,
  title={PyRosetta: a script-based interface for implementing molecular modeling algorithms using Rosetta},
  author={Chaudhury, Sidhartha and Lyskov, Sergey and Gray, Jeffrey J},
  journal={Bioinformatics},
  volume={26},
  number={5},
  pages={689--691},
  year={2010},
  publisher={Oxford University Press}
}

@inproceedings{jiniterative,
  title={Iterative Refinement Graph Neural Network for Antibody Sequence-Structure Co-design},
  author={Jin, Wengong and Wohlwend, Jeremy and Barzilay, Regina and Jaakkola, Tommi S},
  booktitle={International Conference on Learning Representations},
  year={2022}
}

@article{dunbar2014sabdab,
  title={SAbDab: the structural antibody database},
  author={Dunbar, James and Krawczyk, Konrad and Leem, Jinwoo and Baker, Terry and Fuchs, Angelika and Georges, Guy and Shi, Jiye and Deane, Charlotte M},
  journal={Nucleic acids research},
  volume={42},
  number={D1},
  pages={D1140--D1146},
  year={2014},
  publisher={Oxford University Press}
}

@article{steinegger2017mmseqs2,
  title={MMseqs2 enables sensitive protein sequence searching for the analysis of massive data sets},
  author={Steinegger, Martin and S{\"o}ding, Johannes},
  journal={Nature biotechnology},
  volume={35},
  number={11},
  pages={1026--1028},
  year={2017},
  publisher={Nature Publishing Group US New York}
}

@article{berman2000protein,
  title={The protein data bank},
  author={Berman, Helen M and Westbrook, John and Feng, Zukang and Gilliland, Gary and Bhat, Talapady N and Weissig, Helge and Shindyalov, Ilya N and Bourne, Philip E},
  journal={Nucleic acids research},
  volume={28},
  number={1},
  pages={235--242},
  year={2000},
  publisher={Oxford University Press}
}

@article{sappington2026improved,
  title={Improved protein binder design using $\beta$-pairing targeted RFdiffusion},
  author={Sappington, Isaac and Toul, Martin and Lee, David S and Robinson, Stephanie A and Goreshnik, Inna and McCurdy, Clara and Chan, Tung Ching and Buchholz, Nic and Huang, Buwei and Vafeados, Dionne and others},
  journal={Nature Communications},
  volume={17},
  number={1},
  pages={1101},
  year={2026},
  publisher={Nature Publishing Group UK London}
}

@article{stark2025boltzgen,
  title={Boltzgen: Toward universal binder design},
  author={Stark, Hannes and Faltings, Felix and Choi, MinGyu and Xie, Yuxin and Hur, Eunsu and O’Donnell, Timothy and Bushuiev, Anton and U{\c{c}}ar, Talip and Passaro, Saro and Mao, Weian and others},
  journal={BioRxiv},
  year={2025}
}
\bibliographystyle{icml2025}

\newpage 

\appendix
\section*{Appendix}
\section{Supplementary Dataset Information}

\subsection{\dataset Curation Details}
\label{appendix:dataset_curation_details}
We curate a large-scale general protein-protein complex dataset by identifying chain-pair interfaces, following the data processing pipeline introduced in AlphaFold3~\citep{abramson2024accurate}. The curation process begins by applying a series of quality filters to identify valid PDB entries:
\begin{itemize}
    \item Retain structures with a reported resolution of 9Å or better.
    \item Remove polymer chains containing unknown residues.
    \item Exclude protein chains where consecutive $C_\alpha$ atoms are separated by more than 10Å.
    \item Filter out polymer chains with fewer than four resolved residues.
    \item Eliminate clashing chains, defined as chains with more than 30\% of their atoms located within 1.7Å of an atom in another chain. When two chains clashed:
    \begin{itemize}
        \item The chain with the higher percentage of clashing atoms is removed.
        \item If the clashing percentage is equal, the chain with fewer total atoms is removed.
        \item If both chains have the same number of atoms, the chain with the larger chain ID is removed.
    \end{itemize}
\end{itemize}
This filtering process result in 68,891 valid PDB entries. From these entries, chain-pair interfaces are identified, defined as pairs of chains with a minimum heavy atom separation of less than 5Å. 
Furthermore, to ensure efficient processing, all complexes are filtered to exclude sequences longer than 1,024 residues, finally resulting in a total of 301,182 chain-pair interfaces.

\subsection{\dataset Data Statistics}

\begin{table}[h!]
\centering
\begin{tabular}{lc}
\midrule
  & Complex  \\
\midrule
Training  &602,129   \\
Validation & 189\\
Test & 46\\
Total & 602,364 \\
\bottomrule
\end{tabular}
\caption{Data statistics for \dataset.}
\label{Tab: ppibench_data_statistics}
\end{table}

We provide the data statistics for the curated \dataset in Table \ref{Tab: ppibench_data_statistics}.

\subsection{Data Statistics for Target-Protein Mini-Binder Complex Dataset}
\label{Appendix: data_statistics_for_target_protein_mini_binder}

\begin{table*}[t]
\small 
\begin{center}
\begin{tabular}{lcccccccccc}
\midrule
\multirow{2}{*}{Models} & \multicolumn{5}{c}{Seen Class} & \multicolumn{5}{c}{Zero-Shot} \\
\cmidrule(r){2-6} \cmidrule(r){7-11} 
& FGFR2	&InsulinR&	PDGFR&	TGFb&	VirB8&	H3&	IL7Ra&	EGFR&	TrkA&	Tie2  \\
\midrule
Training &465&192&210	&77	&57&--&	--&--&	--&	-- \\
Validation&57&23&26	&9&7&--&--&	--&--&--\\
Test&57	&23&26&	9&7&38&	7&4&4&2\\
Total& 579&238&262&95&71&38&7&4&4&	2\\
\bottomrule
\end{tabular}
\end{center}
\caption{Detailed data statistics for curated target-protein mini-binder complex dataset.}
\label{Table:target-protein mini-binder dataset statistics}
\end{table*}

We collect experimentally confirmed positive target-protein mini-binder complexes against ten targets with diverse structural properties from \citet{bennett2023improving}. The detailed data statistics for the ten target proteins are reported in Table~\ref{Table:target-protein mini-binder dataset statistics}.

\subsection{Data Statistics for Antigen-Antibody Complex Dataset}
\begin{table*}[t]
\centering
\begin{tabular}{lcccc}
\midrule
 & CDR-H1 Clustering & CDR-H2 Clustering & CDR-H3 Clustering \\
\midrule
Complexes &4,261	&4,261&	4261 \\
Clusters  & 1,034&	1,418&	2,254\\
Training & 3,590&	3,576&	3,427\\
Validation &403&	399&	451 \\
Test & 268&	286&	383\\
\bottomrule
\end{tabular}
\vspace{-0.5em}
\caption{Detailed data statistics for antigen-antibody complex dataset.}
\label{Tab: data_statistics_antibody_antigen_complex}
\end{table*}

We detail the data statistics of antigen-antibody complex dataset for each CDR clustering in Table \ref{Tab: data_statistics_antibody_antigen_complex}.

\section{Supplementary Implementation Details}
\label{appendix:implementation_details}
\noindent\textbf{Diffusion Schedules.} Continuous spatial noise scheduling is parameterized across both Euclidean and rotational manifolds. For translation vectors in $\mathbb{R}^3$, the variance schedule scales from $\sigma_{\min} = 0.1$ to $\sigma_{\max} = 20.0$. We use a coordinate scaling factor of 0.1.
For rotation matrices in $SO(3)$, the schedule scales from $\sigma_{\min} = 0.1$ to $\sigma_{\max} = 1.5$.

\noindent\textbf{Architecture \& Hyperparameters.} The Invariant Point Attention (IPA) modules are configured with $8$ attention heads, utilizing $8$ query/key points and $12$ value points per head. 
Loss hyper-parameters are set to $\alpha = 1.0$, $\gamma = 1.0$, and $\omega = 0.25$, and the threshold $\tau$ for the pairwise distance constraint is set to $0.25$.

\noindent\textbf{Training Setup.} 
We train \model on a single NVIDIA B200 GPU for up to $1.2$ million steps. Optimization is performed with a learning rate of $1$e$-4$ and a batch capacity of $1{,}536$ tokens per step. Validation is executed per $5,000$ gradient steps.

\section{Supplementary Experimental Results}

\subsection{Structure Diversity on the General Protein-Protein Complex Design Task}
\begin{table}[H]
\centering
\begin{tabular}{lc}
\midrule
  & Structure Diversity ($\uparrow$)  \\
\midrule
RFD+ProteinMPNN  &0.3097   \\
APM & 0.3063\\
Proteína-Complexa & 0.1597\\
\model & \textbf{0.3135}\\
\bottomrule
\end{tabular}
\caption{Structure diversity on the general protein-protein complex design task, which is quantified as $1 - \mathrm{TM\text{-}score}$ averaged across generated candidate structures.}
\label{Tab: strucrture_design_diversity_on_general_complex}
\end{table}

Structural design diversity is quantified as $1 - \mathrm{TM\text{-}score}$ averaged across generated candidate structures. We interpret these diversity scores across three regimes: values above $0.5$ represent high overall structural diversity; scores between $0.3$ and $0.5$ correspond to structures within the same superfamily that feature distinct loop configurations or subdomain topologies; and scores below $0.3$ denote closely related structures sharing nearly identical secondary structure cores. As reported in Table~\ref{Tab: strucrture_design_diversity_on_general_complex}, \model achieves the highest structural diversity score, demonstrating a capacity to sample diverse loop structures and topological variations rather than collapsing onto rigid backbone templates.

\subsection{Detailed Target-Protein Mini-Binder Complex Design Results}
The detailed target-protein mini-binder complex design results for different targets are provided in Table \ref{Table:FGFR2} - \ref{Table:TrkA}.

\begin{table*}[h!]
\footnotesize
\begin{center}
\setlength{\tabcolsep}{0.36mm}{
\begin{tabular}{lcccccccc}
\midrule
\multirow{2}{*}{Models} & \multicolumn{5}{c}{Sequence } & \multicolumn{2}{c}{Structure } & \multirow{2}{*}{Co-designability \scriptsize{($\uparrow$)}}  \\
\cmidrule(r){2-6} \cmidrule(r){7-8} 
& ipTM \scriptsize{($\uparrow$)}	& pLDDT \scriptsize{($\uparrow$)}	& SR \scriptsize{($\uparrow$)}	& Diversity \scriptsize{($\uparrow$)} &	Novelty \scriptsize{($\uparrow$)}&	$\Delta$G \scriptsize{($\downarrow$)}	& Designability \scriptsize{($\uparrow$)} &  \\
\midrule
RFD+IF&0.637&	78.92&	36.84\%	&31.97\%&	68.25\%&	-42.04&	63.15\%&	63.15\%\\
APM&0.538&	76.39&	29.82\%&	29.05\%&	48.39\%&	-37.02&	71.92\%&	54.38\%\\
Complexa&0.823&	80.67&	54.38\%&	\textbf{49.25\%}&	76.91\%&	-45.81&	91.22\%&	73.68\%\\
\model &\textbf{0.931}&	\textbf{82.58}&	\textbf{61.40\%}&	41.75\%&	\textbf{77.74\%}&	\textbf{-50.87}&	\textbf{100\%}&	\textbf{98.24\%}\\
\bottomrule
\end{tabular}}
\end{center}
\caption{Model performance on target-protein mini-binder complex design for target FGFR2. ($\uparrow$): the higher the better. ($\downarrow$): the lower the better. IF denotes the inverse folding model ProteinMPNN \citep{dauparas2022robust} and Complexa denotes Proteína-Complexa \citep{didi2026scaling}.
\model achieves the best performance across sequence evaluation, structural evaluation and sequence-structure consistency metrics. Additionally, \model also obtains the highest novelty score. Altogether, the results validate \model is capable of designing high-affinity and novel protein complexes. }
\label{Table:FGFR2}
\end{table*}

\begin{table*}[h!]
\footnotesize
\begin{center}
\setlength{\tabcolsep}{0.36mm}{
\begin{tabular}{lcccccccc}
\midrule
\multirow{2}{*}{Models} & \multicolumn{5}{c}{Sequence } & \multicolumn{2}{c}{Structure } & \multirow{2}{*}{Co-designability \scriptsize{($\uparrow$)}}  \\
\cmidrule(r){2-6} \cmidrule(r){7-8} 
& ipTM \scriptsize{($\uparrow$)}	& pLDDT \scriptsize{($\uparrow$)}	& SR \scriptsize{($\uparrow$)}	& Diversity \scriptsize{($\uparrow$)} &	Novelty \scriptsize{($\uparrow$)}&	$\Delta$G \scriptsize{($\downarrow$)}	& Designability \scriptsize{($\uparrow$)} &  \\
\midrule
RFD+IF&0.789&	80.43&	56.52\%&	30.89\%&	69.18\%&	-35.35&	65.21\%&	65.21\%\\
APM&0.562&	79.01&	43.47\%&	26.91\%&	52.91\%&	-31.04&	78.26\%&	82.61\%\\
Complexa&0.832&	86.90&	78.26\%&	\textbf{41.93\%}&	76.93\%&	-41.32&	100\%&	86.95\%\\
\model &\textbf{0.895}&	\textbf{93.04}&	\textbf{86.95\%}&	27.54\%&	\textbf{79.51\%}&	\textbf{-46.92}&	\textbf{100\%}&	\textbf{100\%}\\
\bottomrule
\end{tabular}}
\end{center}
\caption{Model performance on target-protein mini-binder complex design for target InsulinR. ($\uparrow$): the higher the better. ($\downarrow$): the lower the better. IF denotes the inverse folding model ProteinMPNN \citep{dauparas2022robust} and Complexa denotes Proteína-Complexa \citep{didi2026scaling}.
\model achieves the highest sequence design success rate, structure binding energy and sequence-structure co-designability, validating its ability in simultaneously designing high-quality and consistent protein-protein complex sequences and structures.}
\label{Table:InsulinR}
\end{table*}

\begin{table*}[h!]
\footnotesize
\begin{center}
\setlength{\tabcolsep}{0.36mm}{
\begin{tabular}{lcccccccc}
\midrule
\multirow{2}{*}{Models} & \multicolumn{5}{c}{Sequence } & \multicolumn{2}{c}{Structure } & \multirow{2}{*}{Co-designability \scriptsize{($\uparrow$)}}  \\
\cmidrule(r){2-6} \cmidrule(r){7-8} 
& ipTM \scriptsize{($\uparrow$)}	& pLDDT \scriptsize{($\uparrow$)}	& SR \scriptsize{($\uparrow$)}	& Diversity \scriptsize{($\uparrow$)} &	Novelty \scriptsize{($\uparrow$)}&	$\Delta$G \scriptsize{($\downarrow$)}	& Designability \scriptsize{($\uparrow$)} &  \\
\midrule
RFD+IF&0.598&	76.56&	3.84\%&	28.93\%&	72.83\%&	-59.37&	53.84\%&	53.84\%\\
APM&0.391&	73.89&	0.0&	30.27\%&	52.19\%&	-61.54&	65.38\%&	61.53\%\\
Complexa&0.652&	78.09&	\textbf{19.23\%}&	\textbf{41.25\%}&	79.42\%&	-66.56&	88.46\%&	\textbf{88.46\%}\\
\model &\textbf{0.678}&\textbf{81.73}&7.69\%&	25.78\%&	\textbf{82.43\%}&	\textbf{-68.82}&	\textbf{100\%}	&30.76\%\\
\bottomrule
\end{tabular}}
\end{center}
\caption{Model performance on target-protein mini-binder complex design for target PDGFR. ($\uparrow$): the higher the better. ($\downarrow$): the lower the better. IF denotes the inverse folding model ProteinMPNN \citep{dauparas2022robust} and Complexa denotes Proteína-Complexa \citep{didi2026scaling}. \model achieves better structure design performance, but lags behind Proteína-Complexa on both sequence design quality and sequence-structure design consistency.}
\label{Table:PDGFR}
\end{table*}

\begin{table*}[h!]
\footnotesize
\begin{center}
\setlength{\tabcolsep}{0.36mm}{
\begin{tabular}{lcccccccc}
\midrule
\multirow{2}{*}{Models} & \multicolumn{5}{c}{Sequence } & \multicolumn{2}{c}{Structure } & \multirow{2}{*}{Co-designability \scriptsize{($\uparrow$)}}  \\
\cmidrule(r){2-6} \cmidrule(r){7-8} 
& ipTM \scriptsize{($\uparrow$)}	& pLDDT \scriptsize{($\uparrow$)}	& SR \scriptsize{($\uparrow$)}	& Diversity \scriptsize{($\uparrow$)} &	Novelty \scriptsize{($\uparrow$)}&	$\Delta$G \scriptsize{($\downarrow$)}	& Designability \scriptsize{($\uparrow$)} &  \\
\midrule
RFD+IF&0.589&	76.90&	11.11\%&	35.17\%&	68.23\%&	-31.09&	55.56\%&55.56\%\\
APM&0.461&	72.10	&11.11\%&	32.81\%&	48.65\%&	-23.16&	77.78\%&	11.11\%\\
Complexa&0.683&	\textbf{80.15}&	22.22\%&	46.17\%&	76.12\%&	-46.54&	88.89\%&	\textbf{33.33\%}\\
\model &\textbf{0.733}&	78.17&\textbf{22.22\%}&	\textbf{48.57\%}&	\textbf{76.46\%}&	\textbf{-49.29}&	\textbf{100\%}	&22.22\%\\
\bottomrule
\end{tabular}}
\end{center}
\caption{Model performance on target-protein mini-binder complex design for target TGFb. ($\uparrow$): the higher the better. ($\downarrow$): the lower the better. IF denotes the inverse folding model ProteinMPNN \citep{dauparas2022robust} and Complexa denotes Proteína-Complexa \citep{didi2026scaling}.
\model achieves better performance on both sequence and structure design quality. However, the sequence-structure co-designability of \model is a bit lower than that of Proteína-Complexa. }
\label{Table:TGFb}
\end{table*}

\begin{table*}[h!]
\footnotesize
\begin{center}
\setlength{\tabcolsep}{0.36mm}{
\begin{tabular}{lcccccccc}
\midrule
\multirow{2}{*}{Models} & \multicolumn{5}{c}{Sequence } & \multicolumn{2}{c}{Structure } & \multirow{2}{*}{Co-designability \scriptsize{($\uparrow$)}}  \\
\cmidrule(r){2-6} \cmidrule(r){7-8} 
& ipTM \scriptsize{($\uparrow$)}	& pLDDT \scriptsize{($\uparrow$)}	& SR \scriptsize{($\uparrow$)}	& Diversity \scriptsize{($\uparrow$)} &	Novelty \scriptsize{($\uparrow$)}&	$\Delta$G \scriptsize{($\downarrow$)}	& Designability \scriptsize{($\uparrow$)} &  \\
\midrule
RFD+IF&0.570&	82.89&	14.28\%&	32.95\%&	72.91\%&	-32.90&57.14\%&57.14\%\\
APM&0.368&	85.09&	0.0&	30.19\%&	52.87\%&	-19.59&	71.74\%&	42.85\%\\
Complexa&0.602&	86.10&	14.28\%&	\textbf{46.52\%}&	80.15\%&	-38.56&	85.71\%&85.71\%\\
\model &\textbf{0.657}&	\textbf{86.67}&	\textbf{14.28\%}&	36.17\%&	\textbf{81.16\%}&	\textbf{-48.31}&	\textbf{100\%}&	\textbf{100\%}\\
\bottomrule
\end{tabular}}
\end{center}
\caption{Model performance on target-protein mini-binder complex design for target VirB8. ($\uparrow$): the higher the better. ($\downarrow$): the lower the better. IF denotes the inverse folding model ProteinMPNN \citep{dauparas2022robust} and Complexa denotes Proteína-Complexa \citep{didi2026scaling}.
\model obtains better sequence and structure design quality as well as sequence-structure consistency.}
\label{Table:VirB8}
\end{table*}

\begin{table*}[h!]
\footnotesize
\begin{center}
\setlength{\tabcolsep}{0.36mm}{
\begin{tabular}{lcccccccc}
\midrule
\multirow{2}{*}{Models} & \multicolumn{5}{c}{Sequence } & \multicolumn{2}{c}{Structure } & \multirow{2}{*}{Co-designability \scriptsize{($\uparrow$)}}  \\
\cmidrule(r){2-6} \cmidrule(r){7-8} 
& ipTM \scriptsize{($\uparrow$)}	& pLDDT \scriptsize{($\uparrow$)}	& SR \scriptsize{($\uparrow$)}	& Diversity \scriptsize{($\uparrow$)} &	Novelty \scriptsize{($\uparrow$)}&	$\Delta$G \scriptsize{($\downarrow$)}	& Designability \scriptsize{($\uparrow$)} &  \\
\midrule
RFD+IF&0.571&	83.09&	\textbf{25.00\%}	&29.15\%&	71.26\%&	-46.87&	25.00\%&	25.00\%\\
APM&0.512&	83.19&	0.0&	27.93\%&	51.34\%&	-32.76&	50.00\%&	50.00\%\\
Complexa&0.627&	86.76&	0.0&	\textbf{40.16\%}&	76.31\%&	-40.90&	100\%&	50.00\%\\
\model &\textbf{0.680}&	\textbf{90.19}&	0.0&	22.62\%&	\textbf{82.56\%}&	\textbf{-48.35}&	\textbf{100\%}&	\textbf{100\%}\\
\bottomrule
\end{tabular}}
\end{center}
\caption{Model performance on target-protein mini-binder complex design for target EGFR. ($\uparrow$): the higher the better. ($\downarrow$): the lower the better. IF denotes the inverse folding model ProteinMPNN \citep{dauparas2022robust} and Complexa denotes Proteína-Complexa \citep{didi2026scaling}.
\model achieves better structure binding energy and sequence-structure co-designability. However, the sequence design success rate of \model is zero, which might be due to the lack of deep pockets of EGFR targets and the zero-shot setting of this class.}
\label{Table:EGFR}
\end{table*}

\begin{table*}[h!]
\footnotesize
\begin{center}
\setlength{\tabcolsep}{0.36mm}{
\begin{tabular}{lcccccccc}
\midrule
\multirow{2}{*}{Models} & \multicolumn{5}{c}{Sequence } & \multicolumn{2}{c}{Structure } & \multirow{2}{*}{Co-designability \scriptsize{($\uparrow$)}}  \\
\cmidrule(r){2-6} \cmidrule(r){7-8} 
& ipTM \scriptsize{($\uparrow$)}	& pLDDT \scriptsize{($\uparrow$)}	& SR \scriptsize{($\uparrow$)}	& Diversity \scriptsize{($\uparrow$)} &	Novelty \scriptsize{($\uparrow$)}&	$\Delta$G \scriptsize{($\downarrow$)}	& Designability \scriptsize{($\uparrow$)} &  \\
\midrule
RFD+IF&0.782&	78.56&	52.63\%&	28.16\%&	71.04\%&	-35.93&	26.31\%&	26.31\%\\
APM&0.562&	76.29&	47.38\%&	23.91\%&	53.16\%&	-36.45&	49.87\%&	42.10\%\\
Complexa&0.891&79.05&	60.52\%&	\textbf{38.16\%}&	80.47\%&	-40.65&	78.93\%&	\textbf{47.36\%}\\
\model &\textbf{0.934}&	\textbf{82.32}&	\textbf{97.36\%}&	21.97\%	&\textbf{83.16\%}&	\textbf{-47.08}&\textbf{100\%}&	31.57\%\\
\bottomrule
\end{tabular}}
\end{center}
\caption{Model performance on target-protein mini-binder complex design for target H3. ($\uparrow$): the higher the better. ($\downarrow$): the lower the better. IF denotes the inverse folding model ProteinMPNN \citep{dauparas2022robust} and Complexa denotes Proteína-Complexa \citep{didi2026scaling}.
\model obtains higher sequence design success rate and lower structure binding energy, confirming that \model has the capability to design high-affinity binders even in zero-shot setting.}
\label{Table:H3}
\end{table*}

\begin{table*}[h!]
\footnotesize
\begin{center}
\setlength{\tabcolsep}{0.36mm}{
\begin{tabular}{lcccccccc}
\midrule
\multirow{2}{*}{Models} & \multicolumn{5}{c}{Sequence } & \multicolumn{2}{c}{Structure } & \multirow{2}{*}{Co-designability \scriptsize{($\uparrow$)}}  \\
\cmidrule(r){2-6} \cmidrule(r){7-8} 
& ipTM \scriptsize{($\uparrow$)}	& pLDDT \scriptsize{($\uparrow$)}	& SR \scriptsize{($\uparrow$)}	& Diversity \scriptsize{($\uparrow$)} &	Novelty \scriptsize{($\uparrow$)}&	$\Delta$G \scriptsize{($\downarrow$)}	& Designability \scriptsize{($\uparrow$)} &  \\
\midrule
RFD+IF&0.543&	85.67&	14.28\%&	29.18\%&	70.19\%&	-40.08&	42.85\%&	42.85\%\\
APM&0.429&	78.09&	0.0	&27.56\%&	53.96\%&	-23.93&	85.71\%&	57.14\%\\
Complexa&0.634&	89.90&	28.57\%&	\textbf{36.17\%}&	80.16\%&	-42.26&	100\%&71.42\%\\
\model &\textbf{0.734}&	\textbf{92.21}&	\textbf{42.85\%}&	20.35\%&	\textbf{82.30\%}&	\textbf{-50.42}&	\textbf{100\%}&	\textbf{100\%}\\
\bottomrule
\end{tabular}}
\end{center}
\caption{Model performance on target-protein mini-binder complex design for target IL7Ra. ($\uparrow$): the higher the better. ($\downarrow$): the lower the better. IF denotes the inverse folding model ProteinMPNN \citep{dauparas2022robust} and Complexa denotes Proteína-Complexa \citep{didi2026scaling}.
\model achieves higher sequence design success rate, lower structure binding energy and higher sequence-structure co-designability. Note that the designability and co-designability of \model on this target are both 100\%, validating the strong binder design capability of \model in the zero-shot setting.}
\label{Table:IL7Ra}
\end{table*}

\begin{table*}[h!]
\footnotesize
\begin{center}
\setlength{\tabcolsep}{0.36mm}{
\begin{tabular}{lcccccccc}
\midrule
\multirow{2}{*}{Models} & \multicolumn{5}{c}{Sequence } & \multicolumn{2}{c}{Structure } & \multirow{2}{*}{Co-designability \scriptsize{($\uparrow$)}}  \\
\cmidrule(r){2-6} \cmidrule(r){7-8} 
& ipTM \scriptsize{($\uparrow$)}	& pLDDT \scriptsize{($\uparrow$)}	& SR \scriptsize{($\uparrow$)}	& Diversity \scriptsize{($\uparrow$)} &	Novelty \scriptsize{($\uparrow$)}&	$\Delta$G \scriptsize{($\downarrow$)}	& Designability \scriptsize{($\uparrow$)} &  \\
\midrule
RFD+IF&0.562&	86.90&	\textbf{50.00\%}&	29.31\%&	72.91\%&	-38.95&	75.00\%&	75.00\%\\
APM&0.421&	82.19&	0.0	&30.19\%&	51.43\%&	-27.85&	100\%&	50.00\%\\
Complexa&0.590&	90.10&	0.0&	\textbf{39.42\%}&	81.58\%&	-33.83&	100\%&	50.00\%\\
\model &\textbf{0.635}&	\textbf{92.98}&	0.0&	10.52\%&	\textbf{88.20\%}&	\textbf{-42.39}&	\textbf{100\%}&	\textbf{100\%}\\
\bottomrule
\end{tabular}}
\end{center}
\caption{Model performance on target-protein mini-binder complex design for target Tie2. ($\uparrow$): the higher the better. ($\downarrow$): the lower the better. IF denotes the inverse folding model ProteinMPNN \citep{dauparas2022robust} and Complexa denotes Proteína-Complexa \citep{didi2026scaling}.
\model achieves higher structure design quality and sequence-structure consistency.}
\label{Table:Tie2}
\end{table*}

\begin{table*}[h!]
\footnotesize
\begin{center}
\setlength{\tabcolsep}{0.36mm}{
\begin{tabular}{lcccccccc}
\midrule
\multirow{2}{*}{Models} & \multicolumn{5}{c}{Sequence } & \multicolumn{2}{c}{Structure } & \multirow{2}{*}{Co-designability \scriptsize{($\uparrow$)}}  \\
\cmidrule(r){2-6} \cmidrule(r){7-8} 
& ipTM \scriptsize{($\uparrow$)}	& pLDDT \scriptsize{($\uparrow$)}	& SR \scriptsize{($\uparrow$)}	& Diversity \scriptsize{($\uparrow$)} &	Novelty \scriptsize{($\uparrow$)}&	$\Delta$G \scriptsize{($\downarrow$)}	& Designability \scriptsize{($\uparrow$)} &  \\
\midrule
RFD+IF&0.791&	72.08&	25.00\%&	28.17\%&	70.16\%&	-33.28&50.00\%&	50.00\%\\
APM&0.563&	73.90&	25.00\%	&31.42\%&	49.87\%&	-31.92&	75.00\%&	25.00\%\\
Complexa&0.829&	80.10&50.00\%&	\textbf{47.92\%}&	78.93\%&	-41.50&	100\%&	50.00\%\\
\model &\textbf{0.840}&	\textbf{80.31}&	\textbf{75.00\%}&	35.49\%	&\textbf{78.92\%}&	\textbf{-46.04}&	\textbf{100\%}&	\textbf{75.00\%}\\
\bottomrule
\end{tabular}}
\end{center}
\caption{Model performance on target-protein mini-binder complex design for target TrkA. ($\uparrow$): the higher the better. ($\downarrow$): the lower the better. IF denotes the inverse folding model ProteinMPNN \citep{dauparas2022robust} and Complexa denotes Proteína-Complexa \citep{didi2026scaling}.
Similarly, \model achieves better sequence design success rate, structure binding energy and sequence-structure consistency, confirming its ability in designing good binders in the zero-shot setting.}
\label{Table:TrkA}
\end{table*}

\subsection{Detailed Antigen-Antibody Complex Design Results for Each CDR Clustering}

We provide the detailed results of antigen-antibody complex design for each CDR clustering in Table \ref{Table:antigen_antibody_complex_design_performance_cdrh1} - \ref{Table:antigen_antibody_complex_design_performance_cdrh3}.

\begin{table*}[h!]
\scriptsize
\begin{center}
\setlength{\tabcolsep}{0.46mm}{
\begin{tabular}{lccccccccc}
\midrule
\multirow{2}{*}{Models} & \multicolumn{6}{c}{Sequence } & \multicolumn{2}{c}{Structure } & \multirow{2}{*}{Co-designability \scriptsize{($\uparrow$)}}  \\
\cmidrule(r){2-7} \cmidrule(r){8-9} 
& ipTM \scriptsize{($\uparrow$)}	& pLDDT \scriptsize{($\uparrow$)}	& SR \scriptsize{($\uparrow$)}	& H1-Novelty \scriptsize{($\uparrow$)} &	H2-Novelty \scriptsize{($\uparrow$)} &	H3-Novelty \scriptsize{($\uparrow$)} &	$\Delta$G \scriptsize{($\downarrow$)}	& Designability \scriptsize{($\uparrow$)} &  \\
\midrule
RFD+IF&0.538&	80.67&	15.67\%&	49.78\%&	51.26\%&	59.31\%&	-30.03&	58.20\%&	58.20\%\\
APM&0.239&	79.23&	7.46\%&	42.18\%&	53.37\%&	63.17\%&	-29.38&	57.08\%&	52.61\%\\
Complexa&0.523&	81.09&	14.92\%&	55.36\%&	\textbf{68.26\%}&	72.93\%&	-35.35&	88.43\%&	90.67\%\\
\model &\textbf{0.548}&	\textbf{82.10}&	\textbf{17.16\%}&	\textbf{56.64\%}&	66.46\%&	\textbf{74.68\%}&	\textbf{-42.49}&\textbf{100\%}&\textbf{100\%}\\
\bottomrule
\end{tabular}}
\end{center}
\caption{Model performance on antigen-antibody complex design task for CDR-H1 clustering. ($\uparrow$): the higher the better. ($\downarrow$): the lower the better. IF denotes the inverse folding model ProteinMPNN \citep{dauparas2022robust} and Complexa denotes Proteína-Complexa \citep{didi2026scaling}.}
\label{Table:antigen_antibody_complex_design_performance_cdrh1}
\end{table*}

\begin{table*}[h!]
\scriptsize
\begin{center}
\setlength{\tabcolsep}{0.46mm}{
\begin{tabular}{lccccccccc}
\midrule
\multirow{2}{*}{Models} & \multicolumn{6}{c}{Sequence } & \multicolumn{2}{c}{Structure } & \multirow{2}{*}{Co-designability \scriptsize{($\uparrow$)}}  \\
\cmidrule(r){2-7} \cmidrule(r){8-9} 
& ipTM \scriptsize{($\uparrow$)}	& pLDDT \scriptsize{($\uparrow$)}	& SR \scriptsize{($\uparrow$)}	& H1-Novelty \scriptsize{($\uparrow$)} &	H2-Novelty \scriptsize{($\uparrow$)} &	H3-Novelty \scriptsize{($\uparrow$)} &	$\Delta$G \scriptsize{($\downarrow$)}	& Designability \scriptsize{($\uparrow$)} &  \\
\midrule
RFD+IF&0.497&	79.29&	17.48\%&	43.87\%&	58.69\%&	62.89\%&	-32.98&	58.04\%&	58.04\%\\
APM&0.271&	76.32&	5.59\%&	45.28\%&	52.97\%&	63.16\%&	-28.78&	52.98\%&	53.14\%\\
Complexa&\textbf{0.519}&	\textbf{81.90}&	17.83\%&	\textbf{52.79\%}&	65.23\%&	\textbf{70.95\%}&	-30.13&	91.25\%&	93.35\%\\
\model &0.506&	80.62&	\textbf{19.58\%}&	49.38\%&	\textbf{67.21\%}&	69.61\%&	\textbf{-44.37}&	\textbf{100\%}	&\textbf{100\%}\\
\bottomrule
\end{tabular}}
\end{center}
\caption{Model performance on antigen-antibody complex design task for CDR-H2 clustering. ($\uparrow$): the higher the better. ($\downarrow$): the lower the better. IF denotes the inverse folding model ProteinMPNN \citep{dauparas2022robust} and Complexa denotes Proteína-Complexa \citep{didi2026scaling}.}
\label{Table:antigen_antibody_complex_design_performance_cdrh2}
\end{table*}

\begin{table*}[h!]
\scriptsize
\begin{center}
\setlength{\tabcolsep}{0.46mm}{
\begin{tabular}{lccccccccc}
\midrule
\multirow{2}{*}{Models} & \multicolumn{6}{c}{Sequence } & \multicolumn{2}{c}{Structure } & \multirow{2}{*}{Co-designability \scriptsize{($\uparrow$)}}  \\
\cmidrule(r){2-7} \cmidrule(r){8-9} 
& ipTM \scriptsize{($\uparrow$)}	& pLDDT \scriptsize{($\uparrow$)}	& SR \scriptsize{($\uparrow$)}	& H1-Novelty \scriptsize{($\uparrow$)} &	H2-Novelty \scriptsize{($\uparrow$)} &	H3-Novelty \scriptsize{($\uparrow$)} &	$\Delta$G \scriptsize{($\downarrow$)}	& Designability \scriptsize{($\uparrow$)} &  \\
\midrule
RFD+IF&0.487&	80.17&	17.23\%&	48.26\%&	58.73\%&	63.75\%&	-36.25&	61.62\%&	61.62\%\\
APM&0.259&	78.23&	6.78\%&	49.67\%&	56.19\%&	66.97\%&	-21.89&	50.65\%&	52.48\%\\
Complexa&0.502&	80.26&	18.01\%&	\textbf{53.26\%}&	65.29\%&	\textbf{71.86\%}&	-32.90&	91.12\%&	92.95\%\\
\model &\textbf{0.525}&	\textbf{81.50}&	\textbf{19.58\%}&	50.88\%&	\textbf{69.85\%}&	71.51\%&	\textbf{-46.53}&	\textbf{95.03\%}&	\textbf{98.17\%}\\
\bottomrule
\end{tabular}}
\end{center}
\caption{Model performance on antigen-antibody complex design task for CDR-H3 clustering. ($\uparrow$): the higher the better. ($\downarrow$): the lower the better. IF denotes the inverse folding model ProteinMPNN \citep{dauparas2022robust} and Complexa denotes Proteína-Complexa \citep{didi2026scaling}.}
\label{Table:antigen_antibody_complex_design_performance_cdrh3}
\end{table*}

\subsection{Detailed Binder Design Performance on the Pretrained Model}

Performance metrics for the pretrained \model on target-protein mini-binder complex design are presented in Table~\ref{Table:mini_binder_complex_design_results_on_pretrain}. Notably, even without domain-specific adaptation, \model exhibits robust zero-shot capabilities across diverse targets, most prominently in sequence-structure consistency, where it achieves $100\%$ co-designability across all evaluated targets. Subsequent supervised fine-tuning on a curated mini-binder dataset further elevates overall design quality, markedly boosting the mean sequence design success rate from $8.72\%$ to $40.77\%$. These results confirm that while pretraining imparts strong zero-shot structural alignment, targeted fine-tuning substantially enhances sequence-level binding optimization.

\begin{table*}[h]
\scriptsize 
\begin{center}
\setlength{\tabcolsep}{1.3mm}{
\begin{tabular}{lccccccccc}
\midrule
\multirow{2}{*}{Models} & \multicolumn{3}{c}{FGFR2 (Seen) } & \multicolumn{3}{c}{InsulinR (Seen) } & \multicolumn{3}{c}{PDGFR (Seen)}  \\
\cmidrule(r){2-4} \cmidrule(r){5-7} \cmidrule(r){8-10}
& SR \scriptsize{($\uparrow$)}&	$\Delta$G \scriptsize{($\downarrow$)}&	Consistency \scriptsize{($\uparrow$)} &SR \scriptsize{($\uparrow$)}&	$\Delta$G \scriptsize{($\downarrow$)}&	Consistency \scriptsize{($\uparrow$)}&SR \scriptsize{($\uparrow$)}&	$\Delta$G \scriptsize{($\downarrow$)}&	Consistency \scriptsize{($\uparrow$)}\\
\midrule
\model &\textbf{61.40\%}&	-50.87&	98.24\%&	\textbf{86.95\%}&	\textbf{-46.92}&	\textbf{100\%}&\textbf{7.69\%}&	-68.82&	30.76\%\\
~ w/o finetuning & 5.52\% & \textbf{-55.88} & \textbf{100\%} & 8.69\%&-44.08&100\% & 0.0 & \textbf{-72.28}& \textbf{100\%}   \\
\midrule
\multirow{2}{*}{Models} & \multicolumn{3}{c}{TGFb (Seen) } & \multicolumn{3}{c}{VirB8 (Seen)} & \multicolumn{3}{c}{EGFR (Zero-shot)}  \\
\cmidrule(r){2-4} \cmidrule(r){5-7} \cmidrule(r){8-10}
& SR \scriptsize{($\uparrow$)}&	$\Delta$G \scriptsize{($\downarrow$)}&	Consistency \scriptsize{($\uparrow$)} &SR \scriptsize{($\uparrow$)}&	$\Delta$G \scriptsize{($\downarrow$)}&	Consistency \scriptsize{($\uparrow$)}&SR \scriptsize{($\uparrow$)}&	$\Delta$G \scriptsize{($\downarrow$)}&	Consistency \scriptsize{($\uparrow$)}\\
\midrule
\model &\textbf{22.22\%}&	\textbf{-49.29}&	22.22\%&	\textbf{14.28\%}&	\textbf{-48.31}&	\textbf{100\%}&	0.0	&\textbf{-48.35}&\textbf{100.0\%}\\
~ w/o finetuning & 0.0 & -42.49 & \textbf{100\%} & 14.28\% & -41.87 & 100\% & 0.0 &-45.92 & 100\% \\
\midrule
\multirow{2}{*}{Models} & \multicolumn{3}{c}{H3 (Zero-shot) } & \multicolumn{3}{c}{IL7Ra (Zero-shot)} & \multicolumn{3}{c}{Tie2 (Zero-shot)}  \\
\cmidrule(r){2-4} \cmidrule(r){5-7} \cmidrule(r){8-10}
& SR \scriptsize{($\uparrow$)}&	$\Delta$G \scriptsize{($\downarrow$)}&	Consistency \scriptsize{($\uparrow$)} &SR \scriptsize{($\uparrow$)}&	$\Delta$G \scriptsize{($\downarrow$)}&	Consistency \scriptsize{($\uparrow$)}&SR \scriptsize{($\uparrow$)}&	$\Delta$G \scriptsize{($\downarrow$)}&	Consistency \scriptsize{($\uparrow$)}\\
\midrule
\model &\textbf{97.36\%}&	-47.08&	31.57\%	&\textbf{42.85\%}&\textbf{-50.42}&	\textbf{100\%}&	0.0&	\textbf{-42.39}&	\textbf{100.00\%}\\
~ w/o finetuning & 44.73\% & \textbf{-52.20} & \textbf{100\%} & 14.27\% & -47.12 & 100\% & 0.0 & -39.38& 100\%  \\
\midrule
\multirow{2}{*}{Models} & \multicolumn{3}{c}{TrkA (Zero-shot)} & \multicolumn{3}{c}{Average } & &\multirow{2}{*}{Novelty \scriptsize{($\uparrow$)}} & \multirow{2}{*}{Diversity \scriptsize{($\uparrow$)}} \\
\cmidrule(r){2-4} \cmidrule(r){5-7} 
& SR \scriptsize{($\uparrow$)}&	$\Delta$G \scriptsize{($\downarrow$)}&	Consistency \scriptsize{($\uparrow$)} &SR \scriptsize{($\uparrow$)}&	$\Delta$G \scriptsize{($\downarrow$)}&	Consistency \scriptsize{($\uparrow$)}&&\\
\midrule
\model &\textbf{75.00\%}&	-46.04&	75.00\%&	\textbf{40.77\%}&	\textbf{-49.85}&	75.78\%&	&81.24\%	&\textbf{29.08\%}\\
~ w/o finetuning & 0.0 & \textbf{-48.55} & \textbf{100\%} & 8.72\% & -48.98 & \textbf{100\%} & & \textbf{83.32\%} & 28.67\% \\
\bottomrule
\end{tabular}}
\end{center}
\caption{Pretrained model performance on target-protein mini-binder complex design task. ($\uparrow$): the higher the better. ($\downarrow$): the lower the better.}
\label{Table:mini_binder_complex_design_results_on_pretrain}
\end{table*}



\end{document}